\documentclass{article} % For LaTeX2e
\usepackage{iclr2024_conference}
\usepackage{times}
\usepackage{graphicx}
\usepackage{amsmath,amsfonts,bm}

\def\eqref#1{equation~\ref{#1}}
\def\1{\bm{1}}

\def\vh{{\bm{h}}}

\DeclareMathAlphabet{\mathsfit}{\encodingdefault}{\sfdefault}{m}{sl}
\SetMathAlphabet{\mathsfit}{bold}{\encodingdefault}{\sfdefault}{bx}{n}

\def\gG{{\mathcal{G}}}

\def\sR{{\mathbb{R}}}

\usepackage{amsmath}
\usepackage{amsthm}
\usepackage{booktabs}
\usepackage{xcolor,colortbl}
\usepackage{tabularx}

\newtheorem{theorem}{Theorem}

\newtheorem{lemma}{Lemma}

\usepackage{graphicx}
\usepackage{multirow}
\usepackage{caption}
\usepackage{wrapfig}
\usepackage{enumitem}
\usepackage{hyperref}
\usepackage{url}
\usepackage[caption=false,farskip=0pt,labelfont={bf}]{subfig}
\newcommand{\msl}{\{\!\!\{}
\newcommand{\msr}{\}\!\!\}}

\title{Neural Common Neighbor with Completion for Link Prediction}

\iclrfinalcopy

\author{
Xiyuan Wang$^{1,2}$\\
\texttt{wangxiyuan@pku.edu.cn}\\
\And
Haotong Yang$^{1,2,3}$\\
\texttt{haotongyang@pku.edu.cn}\\
\And
Muhan Zhang$^{1*}$\\
\texttt{muhan@pku.edu.cn}\\
\AND
$^1$Institute for Artificial Intelligence, Peking University.\\
$^2$School of Intelligence Science and Technology, Peking University.\\
$^3$Key Lab of Machine Perception (MoE)
\thanks{Correspondence to Muhan Zhang.}
}

\begin{document}

\maketitle

\begin{abstract}
In this work, we propose a novel link prediction model and further boost it by studying graph incompleteness. First, we introduce MPNN-then-SF, an innovative architecture leveraging structural feature (SF) to guide MPNN's representation pooling, with its implementation, namely Neural Common Neighbor (NCN). NCN exhibits superior expressiveness and scalability compared with existing models, which can be classified into two categories: SF-then-MPNN, augmenting MPNN's input with SF, and SF-and-MPNN, decoupling SF and MPNN. Second, we investigate the impact of graph incompleteness---the phenomenon that some links are unobserved in the input graph---on SF, like the common neighbor. Through dataset visualization, we observe that incompleteness reduces common neighbors and induces distribution shifts, significantly affecting model performance. To address this issue, we propose to use a link prediction model to complete the common neighbor structure. Combining this method with NCN, we propose Neural Common Neighbor with Completion (NCNC). NCN and NCNC outperform recent strong baselines by large margins, and NCNC further surpasses state-of-the-art models in standard link prediction benchmarks. Our code is available at \url{https://github.com/GraphPKU/NeuralCommonNeighbor}.
\end{abstract}

%\jr{General remarks: This paper first analyzes the limitations of two existing paradigms to incorporate SF into MPNN-based link prediction models, namely SF-then-MPNN and SF-and-MPNN. A new architecture---\emph{MPNN-then-SF} is then proposed, with NCN as its instantiation. Further, the paper analyzes the effects of incompleteness on the common neighbors (and the NCN model), and proposes a method to mitigate such effects. Generally, the motivations are clear, and the text is easy to follow.}

\section{Introduction}
\label{sec::introduction}
\begin{wrapfigure}{r}{0.3\textwidth}
  \begin{center}
  \vskip -0.25in
    \includegraphics[width=0.3\textwidth]{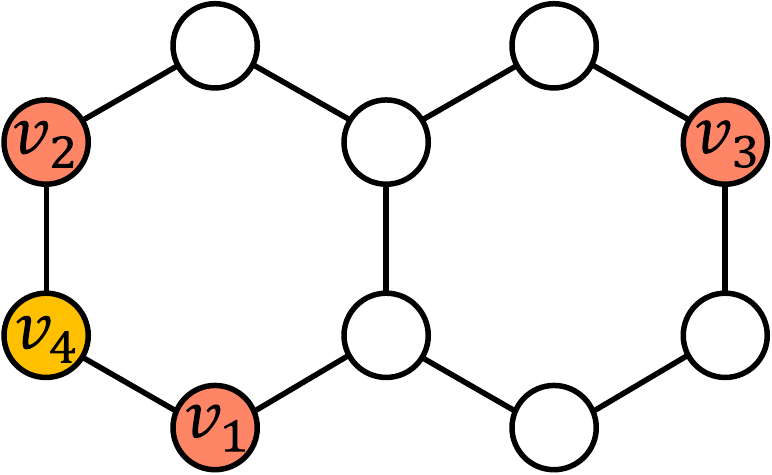} 
  \end{center}
  \vskip -0.15in
  \caption{The failure of MPNN in link prediction task. $v_2$ and $v_3$ have equal MPNN node representations due to symmetry. However, with different pairwise relations, $(v_1, v_2)$ and $(v_1, v_3)$ should have different representations.}\label{fig:GAEfailure}
  \vskip -0.25in
\end{wrapfigure}
Link prediction is a crucial task in graph machine learning, finding applications in various domains, such as recommender systems~\citep{Zhang2020}, knowledge graph completion~\citep{NBFNet}, and drug interaction prediction~\citep{DrugInteraction}. Graph Neural Networks (GNNs) have gained prominence in link prediction tasks, with Graph Autoencoder (GAE)~\citep{GAE} being a notable representation. GAE utilizes Message Passing Neural Network (MPNN)~\citep{MPNN} representations of two individual target nodes to predict link existence. However, \citet{zhang2021labeling} point out a limitation in GAE: it overlooks pairwise relations between target nodes. For example, in Figure~\ref{fig:GAEfailure}, GAE always produces the same prediction for two links $(v_1, v_2)$ and $(v_1, v_3)$  despite their differing pairwise relationships, because MPNN generates identical representations for nodes $v_2, v_3$ due to graph symmetry. Nevertheless, the two links have different structural features. For example, $v_1$ and $v_2$ have a common neighbor $v_4$, while $v_1$ and $v_3$ do not have any. Therefore, various methods combine structural feature (SF) and MPNN for better expressivity and have dominated the link prediction task~\citep{zhang2021labeling,Neo-GNN,Gsketch}. 

\begin{wrapfigure}{r}{7cm}
\centering
\vskip -0.05in
    \includegraphics[width=0.5\textwidth]{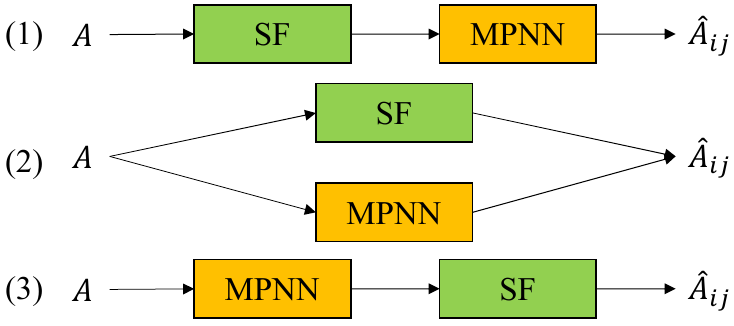}
\vskip -0.05in
    \caption{Archtectures for combining SF and MPNN. $A$ denote the input graph structure. Existing works are (1) SF-then-MPNN (2) SF-and-MPNN architectures. We propose a completely new architecture (3) MPNN-then-SF.}\label{fig:3arch}
\vskip -0.2in
\end{wrapfigure}
However, models combining SF and MPNN still have much room for improvement. They can generally be concluded into two architectures: \textit{SF-then-MPNN} and \textit{SF-and-MPNN}, as shown in Figure~\ref{fig:3arch}. For example, SEAL~\citep{SEAL} adds target-link-specific hand-crafted features to the node features of the input graphs of MPNN, whose output node representations are pooled to represent links.  SEAL belongs to the SF-then-MPNN architecture, using SF to augment the input graph of MPNN. Though SF-then-MPNN models achieve provably high expressivity~\citep{zhang2021labeling}, they require running MPNN on a different graph for each target link, leading to significant computational overhead. In contrast, Neo-GNN~\citep{Neo-GNN} and BUDDY~\citep{Gsketch} decouple the structural feature from MPNN, directly incorporating manually created pairwise features with individual node representations produced by MPNN as link representations, necessitating only a single run of MPNN on the original graph. These models fall under the SF-and-MPNN category, where structural features and MPNN are independent. Such methods have limited expressivity. For example, SEAL can capture the representations of common neighbor nodes, while BUDDY can only count the number of common neighbors. 

To solve the drawback of two architectures above, we propose \textit{MPNN-then-SF} architecture, which initially applies MPNN to the original graph and then uses structural features to guide the pooling of node representations. This approach offers strong expressivity and scalability: similar to SF-then-MPNN models, MPNN-then-SF can capture the node features of common neighbors, and similar to SF-and-MPNN models, it runs MPNN only once for all target links. We introduce the Neural Common Neighbor (NCN) as an instantiation of the MPNN-then-SF architecture. In experiments, NCN outperforms existing models in both scalability and performance.

Furthermore, since NCN heavily relies on common neighbor structure, which is significantly affected by graph incompleteness, we also investigate the impact of incompleteness. Graph incompleteness is ubiquitous in link prediction tasks because the goal is to predict unobserved edges not present in the input graph. We empirically observe that incompleteness reduces the number of common neighbor and leads to a shift in the distribution of common neighbor between the training and test sets. These phenomena collectively lead to performance degradation. To mitigate this issue, we first employ NCN to complete the common neighbor structure and then apply NCN to the completed graph. In experiments, this method significantly improves the performance of NCN.

In conclusion, our contributions are as follows:
\begin{itemize}[itemsep=2pt,topsep=-2pt,parsep=0pt,leftmargin=10pt]
    \item We introduce Neural Common Neighbor (NCN) with MPNN-then-SF architecture for link prediction, demonstrating superior performance and scalability compared to existing models.
    \item We analyze the impact of graph incompleteness and propose Neural Common Neighbor with Completion (NCNC), which completes the input common neighbor structure and applies NCN to the completed graph. NCNC outperforms state-of-the-art models.
\end{itemize}

\section{Preliminary}\label{sec:preliminary}
We consider an undirected graph $\gG\!=\!(V,E,A,X)$, where $V\!=\!\{1,2,\ldots,n\}$ is a set of $n$ nodes, $E \subseteq V\times V$ is the set of edges, $X\in \mathbb R^{n\times F}$ is a node feature matrix whose $v$-th row $X_v$ is the feature of node $v$, and a symmetric adjacency matrix $A \in \mathbb R^{n\times n}$ has $(u, v)$ element $A_uv=1$ if $(u, v) \in E$ and $0$ otherwise. The \textit{degree} of node $u$ is $d(u, A):=\sum_{v=1}^n A_{uv}$. Node $u$'s neighbors are nodes connected to $u$, $N(u, A):=\{v|v\!\in\!V, A_{uv}>0\}$. For simplicity, we use $N(u)$ to denote $N(u, A)$ when $A$ is fixed. \textit{Common neighbor} means nodes connected to both $i$ and $j$: $N(i)\cap N(j)$.

\paragraph{High Order Neighbor.} We define $A^l$ as a high-order adjacency matrix, where $A_{uv}^l$ represents the number of walks of length $l$ between nodes $u$ and $v$. $N(u, A^l)=\{v|v\in V, A^l_{uv}>0\}$ denotes the set of nodes connected to $u$ by a walk of length $l$ in graph $A$. $N_l(u, A)$ denotes the set of nodes whose shortest path distance to $u$ in graph $A$ is $l$. Existing works define \textit{high-order neighbors} as either $N(u, A^l)$ or $N_l(u, A)$. More generally, the neighborhood of $u$ can be expressed as $N_{l_1}(u, A^{l_2})$, returning all nodes with a shortest path distance of $l_1$ to $u$ in the high-order graph $A^{l_2}$. We use $N_{l_1}^{l_2}(u)$ to denote $N_{l_1}(u, A^{l_2})$ when $A$ is fixed. Given a target link $(i, j)$, their general neighborhood overlap is given by $N_{l_1}^{l_2}(i)\cap N_{l_1'}^{l_2'}(j)$, and their neighborhood difference is given by $N_{l_1}^{l_2}(i)\!-\!N_{l_1'}^{l_2'}(j)$.

\paragraph{Message Passing Neural Network (MPNN).}
Comprising message passing layers, MPNN~\citep{MPNN} is a common GNN framework. The $k^{\text{th}}$ layer is as follows.
\begin{equation}
\vh_v^{(k)}=U^{(k)}(\vh_v^{(k-1)},\text{AGG}(\{M^{(k)}(\vh_v^{(k-1)}, \vh_u^{(k-1)})|u\in N(v)\})),
\end{equation}
where $\vh_v^{(k)}$ is the representation of node $v$ at the $k^{\text{th}}$ layer,  $U^{(k)}, M^{(k)}$ are functions like multi-layer perceptron (MLP), and $\text{AGG}$ denotes an aggregation function like sum or max. The initial node representation $\vh_v^{(0)}$ is node feature $X_v$. In each message passing layer, information is aggregated from neighbors to update the node representation. The final node representations produced by MPNN are the output of the last message passing layer, denoted as $\text{MPNN}(v, A, X)=\vh_v^{(K)}$.

% cannot incorporate node features or graph structure information other than the size/degree of (higher-order) common neighbors
\section{Related Work}
\paragraph{Link Prediction Model}\label{sec:heuristic}
There are three primary categories of link prediction models. \textit{Node embeddings}~\citep{Deepwalk,Node2Vec,Line} map each node to an embedding vector and combine target nodes' embeddings to predict link. \textit{Link prediction heuristics}~\citep{liben2003link,CommonNeighbor,RA,AA} use hand-crafted structural features. \textit{GNNs} utilize Graph Neural Networks. Among these GNNs, Graph Autoencoder (GAE)~\citep{GAE} uses the inner product of two target nodes' MPNN representations, $\langle \text{MPNN}(i, A, X), \text{MPNN}(j, A, X) \rangle$, as link $(i, j)$'s representation. It uses MPNN only and fails to capture pairwise relations between nodes. In contrast, various GNNs combining MPNN and structural features~\citep{SEAL,Neo-GNN,Gsketch} have achieved state-of-the-art performance. SEAL~\citep{SEAL} is representative. For a target link $(i, j)$, it initially adds each node's shortest path distance to $(i, j)$ to node feature $X$ and extracts a $k$-hop subgraph, generating augmented node feature $X'$ and adjacency $A'$. Then SEAL applies MPNN to the subgraph and pools node representations within it, $\sum_{u}\text{MPNN}(u, A', X')$, as the target link representation. Other models employ a distinct approach to incorporate structural features. Neo-GNN~\citep{Neo-GNN} and BUDDY~\citep{Gsketch}, for instance, directly apply MPNN to the original graph and concatenate structural features, such as the count of common neighbors, with the Hadamard product of target node MPNN representations, $\text{MPNN}(i, A, X) \odot \text{MPNN}(j, A, X) |\!| \text{structural features}$.
%?? 
\paragraph{Structural Feature}\label{sec:relatedwork:LPmodel}
Structural features for link prediction vary but are generally based on neighborhood overlap. Notable heuristics, such as Common Neighbor (CN), Resource Allocation (RA), and Adamic Adar (AA), use first-order common neighbors to compute scores for a target link $(i,j)$: 
\begin{small}
\begin{equation}\label{equ:heuristics}
\text{CN}(i,j)=\sum_{u\in N\!(\!i\!)\cap N\!(\!j\!)} 1,\quad \text{RA}(i,j)=\sum_{u\in N\!(\!i\!)\cap N\!(\!j\!)} \frac{1}{d(u)},\quad \text{AA}(i,j)=\sum_{u\in N\!(\!i\!)\cap N\!(\!j\!)} \frac{1}{\log d(u)}.
\end{equation}\end{small}

\begin{wraptable}{r}{0.5\textwidth}
\setlength{\tabcolsep}{0.2mm}
\begin{small}
\vskip -0.2in
    \centering
    \caption{Summary of existing models using Equation~(\ref{equ:framework}). FO and HO denote first-order and high-order neighbors, respectively. $\cap$ and $\!-\!$ denote set intersection and difference, respectively.}\label{tab:framework}
\vskip -0.08in
\begin{tabular}{cccccc}
\toprule
 & Model & Neighbor& $\oplus$ & $g(x)$ & $f(x)$ \\ \midrule
\multirow{3}{*}{Heuristics} & CN & FO & $\cap$ & $1$ & $1$ \\
 & RA & FO & $\cap$ & $1$ & $1/x$ \\
 & AA & FO & $\cap$ & $1$ & $1/\log(x)$ \\
 \midrule
\multirow{2}{*}{GNN} & Neo-GNN & HO & $\cap$ & $x$ & MLP \\
 & BUDDY & HO & $\cap\&-$ & $1$ & $1$ \\
% & \textbf{NCN} & FO & $\cap$ & $1$ & \textbf{MPNN}\\
 \bottomrule
\end{tabular}
\end{small}
\vskip -0.15in
\end{wraptable}

Neo-GNN~\citep{Neo-GNN} and BUDDY \citep{Gsketch} extend these heuristics by utilizing higher-order neighbors. Neo-GNN computes features for high-order neighborhood overlap $N^{l_1}(i),N^{l_2}(j)$ as follows,
\begin{equation}
    \sum_{u\in N_1^{l_1}(i)\cap N_1^{l_2}(j)} A^{l_1}_{iu}A^{l_2}_{ju} f(d(u)),
\end{equation}
where $f$ is a learnable function of node degree $d(u)$. BUDDY~\citep{Gsketch} further utilize high-order neighborhood difference. It computes overlap features $\{a_{l_1,l_2}(i,j)| l_1,l_2=1,2,...,k\}$ and difference features $\{b_l(i,j),b_l(j,i)| l=1,2,...,k\}$ as follows:
\begin{small}
\begin{equation}
a_{l_1,l_2}(i,j)= \sum_{u\in N_{l_1}^1(i) \cap  N_{l_2}^1(j)} 1,\quad b_l(i,j) = \sum_{u\in N_{l}^1(i)-\bigcup_{l'=1}^k N_{l'}^1(j)} 1.
\end{equation}
\end{small}
All these pairwise features can be summarized into the following framework. 
\begin{small}
\begin{equation}\label{equ:framework}
\sum_{u\in N_{l_1}^{l_2}(i)\oplus N_{l_1'}^{l_2'}(j)} g(A^{l_2}_{iu})g(A^{l_2'}_{ju}) f(d(u)),
\end{equation}
\end{small}

where $N_{l_1}^{l_2}(i)$ and $N_{l_1'}^{l_2'}(j)$ denote the general neighborhood of $i$ and $j$, $\oplus$ is a set operator like intersection or difference, and $f,g$ are node degree and high-order adjacency weight functions, respectively. Details on how this framework unify existing structure features are shown in Table~\ref{tab:framework}. %\jr{The last sentence ``and $f,g$ ...'' seems out of context.}

\paragraph{Incompleteness of Graph}

Link prediction task is to forecast unobserved edges, inherently making the input graph incomplete. Nevertheless, graph incompleteness can significantly impact structural features, such as common neighbors, and models based on them. This issue has drawn attention in some prior works. \citet{OpenworldKG} examined how unobserved links could distort evaluation scores, with a specific focus on metrics and benchmark design, while our research concentrates on model design. \citet{FakeEdge} explored the consequences of the presence of target links, whereas our emphasis lies in understanding how incompleteness affects common neighbor-based features. Outside the domain of link prediction, \citet{Gcomplete1} and \citet{Gcomplete2} add edges predicted by GAE to the input graph of GNNs. However, their primary objective was node classification tasks, aiming to enhance edges between nodes of the same class while diminishing others. In contrast, our research addresses distribution shifts and information loss stemming from graph incompleteness, offering unique completion methods and insights tailored for link prediction.

\section{Neural Common Neighbor}\label{sec:NCN}

Structural features (SF), such as common neighbors, are commonly employed in link prediction models. Existing approaches combine SF with Message Passing Neural Networks (MPNN) in two manners (illustrated in Figure~\ref{fig:3arch}): SF-then-MPNN and SF-and-MPNN. However, these approaches exhibit limitations in terms of either scalability or expressivity. To address these issues comprehensively, we introduce a novel architecture, MPNN-then-SF, which offers a unique blend of high expressivity and scalability. Subsequently, we present a concrete instantiation of this architecture, Neural Common Neighbor (NCN). All proofs for theorems in this section are in Appendix~\ref{app::proof}.

\subsection{New Architecture Combining MPNN and SF}
\begin{wrapfigure}{r}{4cm}
\vskip -0.5in
    \centering
    \includegraphics[width=0.25\textwidth]{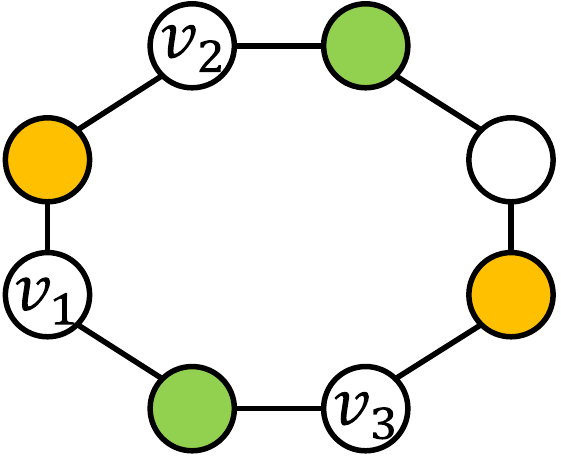}
\vskip -0.05in
    \caption{White, green, and yellow colors represent node features $0, 1$, and $2$, respectively. Both links $(v_1, v_2)$ and $(v_1, v_3)$ have one common neighbor, making it indistinguishable for existing SF-and-MPNN models. However, NCN can differentiate between them because the two common neighbors have different features.}\label{fig::NCNexpexp}
\vskip -0.3in
\end{wrapfigure}
In Figure~\ref{fig:3arch}, we categorize existing methods into two architectures:

\begin{itemize}[itemsep=2pt,topsep=-2pt,parsep=0pt,leftmargin=10pt]
\item SF-then-MPNN: This category includes SEAL~\citep{SEAL} and NBFNet~\citep{NBFNet}. In this approach, the input graph is initially enriched with structural features and then fed into the MPNN, which allows MPNN to leverage SF and have provable expressivity~\citep{SEAL}. However, the drawback is that structural features change with target link, necessitating MPNN to be re-run for each link, resulting in lower scalability.
\iffalse
\begin{equation}
    \text{MPNN}(i, X'(i, j, X, A), A)\odot \text{MPNN}(j, X'(i, j, X, A),  A),
\end{equation}
where $X'(i,j, X, A)$ is the augmented node feature for target link $(i, j)$. $X'$ varies with the model. For SEAL, it is $X$ concatenated with the shortest path distance to $(i, j)$. For NBFNet, it is $X$ concatenated with a onehot vector $\in \sR^{n}$, whose $i$-th element is $1$ and other elements are $0$.
\fi

\item SF-and-MPNN: This category encompasses models like NeoGNN~\citep{Neo-GNN} and BUDDY~\citep{Gsketch}. Here, MPNN takes the original graph as input and runs only once for all target links, leading to high scalability. However, SF are directly concatenated to the final representations and thus detached from MPNN, leading to reduced expressivity.
\iffalse
\begin{equation}
    \text{MPNN}(i, X, A)\odot \text{MPNN}(j, X,  A)|\!| \sum_{u\in N_{l_1}^{l_2}(i)\oplus N_{l_1'}^{l_2'}(j)}g(A^{l_2}_{iu})g(A^{l_2'}_{ju}) f(d(u))
\end{equation}
It usually concatenate GAE representation with structural features directly. 
\fi
\end{itemize}

From these two architectural paradigms, it becomes apparent that feeding the original graph to MPNN is essential for achieving high scalability. Moreover, the coupling between SF and MPNN remains a crucial factor for expressivity. Thus, we introduce a new architecture: MPNN-then-SF. This approach initially runs MPNN on the original graph and then employs structural features to guide the pooling of MPNN features, requiring only one MPNN run and enhancing expressivity. The specific representation of the target link $(i,j)$ is as follows:
\begin{equation}
    \text{Pool}(\{\text{MPNN}(u, A, X)|u\in S\}), 
\end{equation}
where $\text{Pool}$ is a pooling function mapping a multiset of node representations to a single set representation, and $S$ is a node set related to the target link. Multiple node sets can be used in conjunction to produce concatenated representations. This flexible framework can express various models. When using target nodes $i$ and $j$ as $S$ and  Hadamard product as $\text{Pool}$, it can express GAE:
\begin{equation}\label{equ:GenPairwise2}
    \text{MPNN}(i, A, X)\odot \text{MPNN}(j, A, X).
\end{equation}
Alternatively, we can choose $S$ as combinations of high-order neighbor sets of $i$ and $j$, leading to the following form (see Appendix~\ref{app::proof::equ} for the detailed derivation.):
\begin{small}
\begin{equation}\label{equ:GenPairwise3}
\sum_{u\in N_{l_1}^{l_2}(i)\oplus N_{l_1'}^{l_2'}(j)}g(A^{l_2}_{iu})g(A^{l_2'}_{ju})  \text{MPNN}(u, A, X),
\end{equation}
\end{small}

where $g$ is a function transforming the edge weight in high-order adjacency matrix. This framework exhibits stronger expressivity than existing SF-and-MPNN models.
\begin{theorem}\label{thm::expressivity}
Combination of Equation~\ref{equ:GenPairwise2} and Equation~\ref{equ:GenPairwise3} are strictly more expressive than {MPNN-only model: GAE, SF-only models: CN, RA, AA, and MPNN-and-SF models: Neo-GNN, BUDDY.}
\end{theorem}
A key factor contributing to its higher expressivity is the coupling of MPNN and SF. While SF-and-MPNN typically only counts the number of common neighbors, MPNN-then-SF, similar to SF-then-MPNN, can capture node properties of these common neighbors. As shown in Figure~\ref{fig::NCNexpexp}, node tuples $(v_1,v_2)$ and $(v_1,v_3)$ have the same number of common neighbors. However, their common neighbors have different node features, allowing MPNN-then-SF and SF-then-MPNN to distinguish them, a capability that SF-and-MPNN lacks. %$v_2$ and $v_3$ are symmetric and thus have the same MPNN representation, so GAE cannot distinguish $(v_1,v_2)$ and $(v_1,v_3)$. Moreover, $(v_1,v_2)$ and $(v_1,v_3)$ are symmetric if the node feature is ignored, so CN, RA, AA, Neo-GNN, BUDDY cannot distinguish them. However, $(v_1,v_2)$ have a common neighbor with feature $2$, and $(v_1,v_3)$ have a common neighbor with feature $1$, so NCN can distinguish them. 

\subsection{Neural Common Neighbor}

We will now present an implementation for the MPNN-then-SF framework. Notably, the previous models NeoGNN~\citep{Neo-GNN} and BUDDY~\citep{Gsketch} all incorporate higher-order neighbors into their architectures, resulting in significant performance improvements. Surprisingly, in our experiments, we observed that the gains achieved by explicitly considering higher-order neighbors were marginal once we introduced MPNN into the framework (as discussed in Section~\ref{sec::abl}). We speculate that this marginal improvement arises because MPNN implicitly learns information related to higher-order neighbors. Therefore, considering scalability, we opt to utilize only the target nodes and their first-order common neighbors as the node set, leading to the development of our NCN model:
\begin{equation}
\label{equ:NCN}
    \text{NCN}(i,j, A,X)=\text{MPNN}(i, A, X)\odot \text{MPNN}(j, A, X)|\!|\sum_{u\in N(i)\cap N(j)} \text{MPNN}(u, A, X)
\end{equation}
where $g(A_{iu})$ and $g(A_{ju})$ are constants and ignored, and $|\!|$ denotes concatenation. It has high expressivity.
\begin{theorem}\label{thm::expressivity2}
NCN is strictly more expressive than GAE, CN, RA, AA. Moreover, Neo-GNN and BUDDY are not more expressive than NCN.
\end{theorem}
To elaborate, in certain scenarios where the properties of common neighbors hold significant importance, NCN outperforms both BUDDY and Neo-GNN in expressiveness. 

As our first major contribution, NCN represents a straightforward yet potent model for combining structural features and MPNNs. It operates as an implicit high-order model by aggregating first-order common neighbors, each of which implicitly learns higher-order information through MPNN. A comprehensive analysis of time complexity is in Appendix~\ref{app:complexity}.

\begin{figure}[t]
\vskip -0.05in
\centering\subfloat{\includegraphics[width=0.50\textwidth]{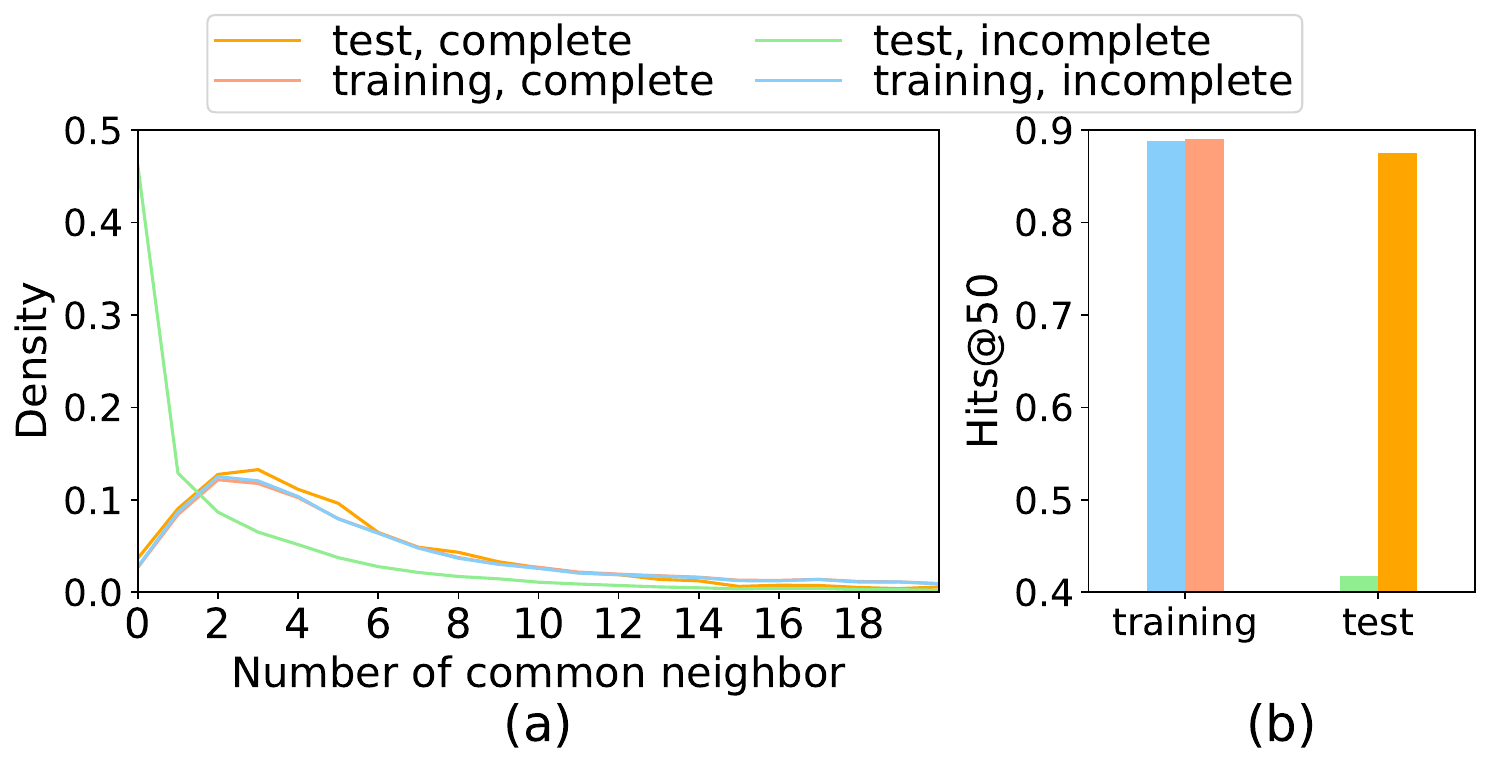}}
   \subfloat{
   \includegraphics[width=0.50\textwidth]{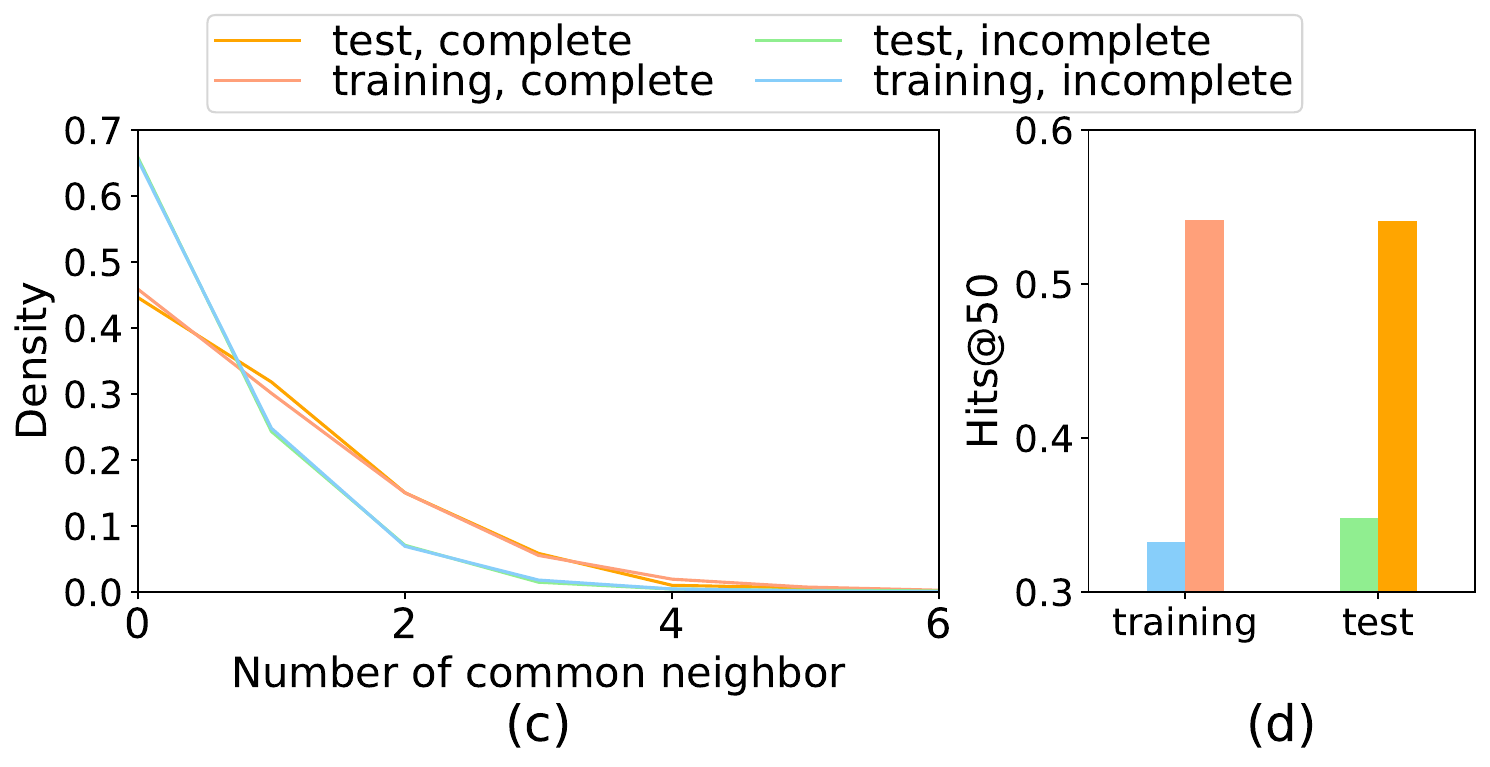}}
    \vskip -0.1in
    \caption{
    Visualization of incompleteness on datasets. The incomplete graph only contains edges in the training set, and the complete graph further contains edges in the validation and test set.
    (a) and (b) visualize the ogbl-collab dataset. (c) and (d) visualize the Cora dataset. (a) and (c) are for distributions of the number of common neighbors of the training edges and test edges. (b) and (d) show performance of CN on the training set and test set. }
    \label{fig:cndist}
    \vskip -0.25in
\end{figure}
\section{Neural Common Neighbor with Completion}\label{sec::intervene}

While NCN outperforms existing models, it relies heavily on the common neighbor structure, which the incompleteness of the graph can notably influence. For instance, in cases where node pairs lack common neighbors, NCN essentially degenerates to GAE, rendering it unable to leverage structural features. Although the absence of common neighbors can suggest that a link is unlikely to exist, certain node pairs may possess common neighbors in the ground truth that remain unobserved in the input graph due to graph incompleteness. Graph incompleteness is ubiquitous in link prediction tasks, given that the objective is to predict unobserved edges. However, few studies have delved into this issue. In this section, we initially demonstrate that incompleteness can result in the loss of common neighbor information, distribution shifts between the training and test sets, and the consequent deterioration of model performance. We propose a straightforward yet effective method to tackle these challenges: common neighbor completion (CNC). CNC completes unobserved common neighbors using a link prediction model. With the introduction of CNC, we enhance NCN and introduce Neural Common Neighbor with Completion (NCNC).

\subsection{Incompleteness Visualization}
\label{sec:incompletenessvis}

To illustrate the challenges posed by incompleteness, we analyze two common datasets: ogbl-collab~\citep{OGB} and Cora~\citep{Cora}. We refer to the graph containing only the edges from the training set as the \textit{incomplete} graph, while the one encompassing edges from the training, validation, and test sets is termed the \textit{complete} graph.

Given the pivotal role of common neighbor information in our NCN model and other link prediction models, we visualize the distribution of the number of common neighbors for training/test edges in both complete and incomplete graphs separately in Figure~\ref{fig:cndist} (a)(c). To assess how incompleteness impacts model performance, we present the performance of CN model (as shown in Section~\ref{sec:relatedwork:LPmodel}) in four distinct scenarios in Figure~\ref{fig:cndist} (b) (d). We observe the following effects of incompleteness:

\paragraph{Loss of Common Neighbors.} Figure~\ref{fig:cndist}(c) illustrates that in the incomplete graph, there are fewer common neighbors for both training and test sets, as indicated by the comparison between the blue (green) and red (orange) lines. When comparing the incomplete and complete graphs, it becomes evident that the incomplete graph suffers from a loss of common neighbor information due to incompleteness. Additionally, more links have no common neighbors at all.

\paragraph{Common Neighbor Distribution Shift.} A noticeable \textit{distribution shift} between the training and test sets is evident in the incomplete graph of the ogbl-collab dataset, as seen in the comparison between the blue and green lines in Figure~\ref{fig:cndist}(a). This shift disappears when the graph is complete (the red and orange lines), indicating that incompleteness is the cause. Such a substantial distribution shift between training and test links could pose challenges in model generalization. This distribution shift is related to the dataset split method. Ogbl-collab is splitted based on the timestamp of edges, and the test edges all belong to the same year. Consequently, compared to the training edges, test edges exhibit stronger correlations with other test edges, resulting in a greater loss of common neighbor when these test edges are absent from the incomplete graph. Conversely, the Cora dataset is randomly splitted, so training and test edges lose a similar ratio of common neighbors and does not exhibit distribution shifts (Figure~\ref{fig:cndist}(c)).

\paragraph{Performance Degradation.} The performance of CN aligns with the common neighbor distribution. In the ogbl-collab dataset, the common neighbor distribution is nearly identical for the training set in both the complete and incomplete graphs, as is the performance (See Figure~\ref{fig:cndist} (b)). However, test performance on the incomplete graph decreases significantly as the test distribution changes. Similar trends are observed in the Cora dataset, with test and training scores declining on incomplete graphs when the common neighbor distribution changes compared to the complete graph.

\paragraph{Remark.} Note that while common neighbor distribution changes may not fully account for the differences between complete and incomplete graphs, they offer valuable insights into how incompleteness alters the input graph structure for other learnable models. %These issues may not manifest in every dataset, and their significance can vary depending on factors such as data type, model design, data split, and more. 
Despite CN is non-learnable and non-generalizable, its calculation for a target edge doesn't involve the edge itself, thereby avoiding data leakage concerns. These findings suggest that having a more complete input graph could yield superior link prediction models. However, in practice, we can only work with the incomplete input graph, necessitating exploring other mitigation methods for these issues.%~\jr{In this paragraph and the previous, it is better to distinguish between ``common neighbor'' and the Common Neighbor link prediction model, e.g. by abbreviating the latter as ``CN''.}

\subsection{Common Neighbor Completion}\label{sec:CNC} 
Motivated by the analysis above, we address graph incompleteness issues with a two-step method:

\paragraph{Soft Completion of Common Neighbors.} We start by softly completing the input graph with a link prediction model, such as NCN. However, instead of completing all edges in the entire graph, which can be impractical for large graphs, we focus specifically on common neighbor links. We compute the probability that a node $u$ serves as a common neighbor for a node tuple $(i, j)$ as follows:
\begin{small}
\begin{equation}\label{equ:completionprob}
P_{uij}=\begin{cases}
    1 &\text{if }u\!\in\!N(i,\!A)\cap N(j,\!A)\\
    \hat A_{iu} &\text{if }u\!\in N(j,\!A)\!-\!N(i,\!A)\\
    \hat A_{ju} &\text{if }u\!\in \! N(i,\!A)\!-\!N(j,\!A)\\
    0&\text{otherwise }
\end{cases}
\end{equation}
\end{small}

where $\hat A_{iu}$ represents the predicted existence probability of link $(i, u)$ by the model. The idea is that $u$ is a common neighbor of $(i, j)$ iff both edges $(i, u)$ and $(j, u)$ exist. If one of these edges is unobserved, we use NCN to predict its link existence probability, which we also use as the probability of $u$ being a common neighbor. In the rare case where both $(i, u)$ and $(j, u)$ are unobserved, we set the probability to $0$. This technique is called "Common Neighbor Completion" (CNC).

\paragraph{Reapplication of NCN on the Completed Graph.}

Following CNC, we apply the NCN model again on the graph that has been completed using the soft common neighbor weights $P_{uij}$. This final model is named \textbf{Neural Common Neighbor with Completion} (NCNC) and is defined as:
\begin{small}
\begin{equation}\label{equ:NCN+C.t}
    \text{NCNC}(i,j,A,X)
    =\text{MPNN}(i, A, X)\odot \text{MPNN}(j, A, X)|\!|\sum_{u\in N(i)\cup N(j)}P_{uij}\text{MPNN}(u, A, X).
\end{equation}
\end{small}

Notably, the input graph of MPNN still takes the original graph as input, allowing it to run only once for all target links, thus maintaining high scalability. While $P_{uij}$ can be predicted using any link prediction model, weak models may not accurately recover the unobserved common neighbor structure. Therefore, in practice, we employ NCN to complete it.

In addition to addressing distribution shifts and common neighbor loss, NCNC also solves the problem that NCN can degrade to GAE when node pairs lack common neighbors. With NCNC, common neighbors are always completed, and the model only degenerates to GAE when both target nodes are isolated nodes. In such cases, where no structural features can be utilized, relying solely on the target node representations is reasonable. For a visual demonstration of CNC's effect, please refer to Appendix~\ref{app:cncexp}, which illustrates how NCN can make more precise predictions by completing common neighbors for node pairs with no observed common neighbors.

\section{Experiment}
\begin{table*}[t]
    \vskip -0.1in
    \centering
    \caption{Results on link prediction benchmarks. The format is average score $\pm$ standard deviation. OOM means out of GPU memory.}\label{tab:main_results}
\vskip -0.05in
\setlength{\tabcolsep}{2pt}
\small{
    \begin{tabular}{lccccccc}
    \toprule
         &
         \textbf{Cora} &  
         \textbf{Citeseer} & 
         \textbf{Pubmed} &
         \textbf{Collab} &
         \textbf{PPA} &
         \textbf{Citation2} 
         &\textbf{DDI} 
         \\
\midrule
          Metric &

          HR@100 &
          HR@100 & 
          HR@100 &
          HR@50 &
          HR@100 &
          MRR 
          &HR@20
         \\ 
         
         \midrule
          
         \textbf{CN} & 
         $33.92 {\scriptstyle \pm 0.46}$& 
         $29.79 {\scriptstyle \pm 0.90}$& 
         $23.13 {\scriptstyle \pm 0.15}$&
         $56.44 {\scriptstyle \pm 0.00}$&
         $27.65 {\scriptstyle \pm 0.00}$&
         $51.47 {\scriptstyle \pm 0.00}$
         &$17.73 {\scriptstyle \pm 0.00}$
         \\

        \textbf{AA} & 
        $39.85 {\scriptstyle \pm 1.34}$&
        $35.19 {\scriptstyle \pm 1.33}$&
        $27.38 {\scriptstyle \pm 0.11}$&
        $64.35 {\scriptstyle \pm 0.00}$&
        $32.45 {\scriptstyle \pm 0.00}$&
        $51.89 {\scriptstyle \pm 0.00}$
        &$18.61 {\scriptstyle \pm 0.00}$
        \\

        \textbf{RA} &
        $41.07 {\scriptstyle \pm 0.48}$ &
        $33.56 {\scriptstyle \pm 0.17}$&
        $27.03 {\scriptstyle \pm 0.35}$& 
        $64.00 {\scriptstyle \pm 0.00}$&
        $49.33{\scriptstyle \pm 0.00}$ & 
        $51.98 {\scriptstyle \pm 0.00}$
        &$27.60 {\scriptstyle \pm 0.00}$
        \\ \midrule
        
\iffalse        
        \textbf{transE} &
        $67.40 {\scriptstyle \pm 1.60}$& 
        $60.19 {\scriptstyle \pm 1.15}$&
        $36.67 {\scriptstyle \pm 0.99}$& 
        $29.40 {\scriptstyle \pm 1.15}$& 
        $22.69 {\scriptstyle \pm 0.49}$ &
        $76.44 {\scriptstyle \pm 0.18}$ 
        & $6.65  {\scriptstyle \pm 0.20}$
        \\
          
        \textbf{complEx} & 
        $37.16 {\scriptstyle \pm 2.76}$& 
        $42.72 {\scriptstyle \pm 1.68}$& 
        $37.80 {\scriptstyle \pm 1.39}$&
        $53.91 {\scriptstyle \pm 0.50}$& 
        $27.42 {\scriptstyle \pm 0.49}$ &
        $72.83 {\scriptstyle \pm 0.38}$ 
        &$8.68 {\scriptstyle \pm 0.36 }$
        \\
        
        \textbf{DistMult} & 
        $41.38 {\scriptstyle \pm 2.49}$& 
        $47.65 {\scriptstyle \pm 1.68}$& 
        $40.32 {\scriptstyle \pm 0.89}$&
        $51.00 {\scriptstyle \pm 0.54}$& 
        $28.61 {\scriptstyle \pm 1.47} $&
        $66.95 {\scriptstyle \pm 0.40} $
        &$11.01 {\scriptstyle \pm 0.49} $
         \\
        \midrule
\fi 
        
        \textbf{GCN} & 
        $66.79 {\scriptstyle \pm 1.65}$& 
        $67.08 {\scriptstyle \pm 2.94}$&
        $53.02 {\scriptstyle \pm 1.39}$& 
        $44.75 {\scriptstyle \pm 1.07}$&
        $18.67 {\scriptstyle \pm 1.32}$&
        $84.74 {\scriptstyle \pm 0.21}$
        &$37.07 {\scriptstyle \pm 5.07}$
         \\
        \textbf{SAGE} & 
        $55.02 {\scriptstyle \pm 4.03}$& 
        $57.01 {\scriptstyle \pm 3.74}$& 
        $39.66 {\scriptstyle \pm 0.72}$& 
        $48.10 {\scriptstyle \pm 0.81}$& 
        $16.55 {\scriptstyle \pm 2.40}$&
        $82.60 {\scriptstyle \pm 0.36}$
        &$53.90 {\scriptstyle \pm 4.74}$ 
        \\ 
        \midrule

        \textbf{SEAL} & 
        $81.71{\scriptstyle \pm 1.30}$& 
        $83.89{\scriptstyle \pm 2.15}$ & 
        $75.54{\scriptstyle \pm 1.32}$&
        $64.74{\scriptstyle \pm 0.43}$& 
        $48.80{\scriptstyle \pm 3.16}$& 
        $87.67{\scriptstyle \pm 0.32}$
        &$30.56{\scriptstyle \pm 3.86}$
        \\ 
        
         \textbf{NBFnet} & 
         $71.65 {\scriptstyle \pm 2.27}$&
         $74.07 {\scriptstyle \pm 1.75}$&
         $58.73 {\scriptstyle \pm 1.99}$&
         OOM&
         OOM&
         OOM
         &$4.00 {\scriptstyle \pm 0.58}$
         \\  
        \midrule
        \textbf{Neo-GNN} & 
        $80.42 {\scriptstyle \pm 1.31} $ &
        $84.67{\scriptstyle \pm 2.16}$ &
        $73.93{\scriptstyle \pm 1.19}$ &
        $57.52 {\scriptstyle \pm 0.37}$& 
        $49.13{\scriptstyle \pm 0.60}$& 
        $87.26{\scriptstyle \pm 0.84}$
        &$63.57{\scriptstyle \pm 3.52}$ 
        \\

\iffalse
         \textbf{ELPH} & 
         $87.72{\scriptstyle \pm 2.13}$&
         $93.44{\scriptstyle \pm 0.53}$&
         $72.99 {\scriptstyle \pm 1.43}$ &
         $66.32{\scriptstyle \pm 0.40}$&
         OOM&
         OOM
         &$83.19{\scriptstyle \pm 2.12}$
         \\ 
\fi         
         \textbf{BUDDY} & 
         $88.00{\scriptstyle \pm 0.44}$&
         $\underline{92.93{\scriptstyle \pm 0.27}}$&
         $74.10{\scriptstyle \pm 0.78}$&
         $\underline{65.94{\scriptstyle \pm 0.58}}$&
         $49.85{\scriptstyle \pm 0.20}$& 
         $87.56{\scriptstyle \pm 0.11}$
         &$78.51{\scriptstyle \pm 1.36}$ 
         \\
         \midrule
         \textbf{NCN} &
         $\underline{89.05{\scriptstyle \pm 0.96}}$&
         ${91.56{\scriptstyle \pm 1.43}}$&
         $\underline{79.05{\scriptstyle \pm 1.16}}$&
         $64.76{\scriptstyle \pm 0.87}$&
         $\underline{61.19{\scriptstyle \pm 0.85}}$& 
         $\underline{88.09{\scriptstyle \pm 0.06}}$
         & $\underline{82.32{\scriptstyle \pm 6.10}}$
         \\
         \textbf{NCNC} &
         $\mathbf{89.65{\scriptstyle \pm 1.36}}$&
         $\mathbf{93.47{\scriptstyle \pm 0.95}}$&
         $\mathbf{81.29{\scriptstyle \pm 0.95}}$&
         $\mathbf{66.61{\scriptstyle \pm 0.71}}$&
         $\mathbf{61.42{\scriptstyle \pm 0.73}}$& 
         $\mathbf{89.12{\scriptstyle \pm 0.40}}$
         &$\mathbf{84.11{\scriptstyle \pm 3.67}}$
         \\ \bottomrule
\end{tabular}
}
\vskip -0.05in
\end{table*}
\begin{table*}[t]
    \centering
\vskip -0.05in
    \caption{Ablation study on link prediction benchmarks. %The format is average score $\pm$ standard deviation. 
    }\label{tab:abl}
\vskip -0.05in
\setlength{\tabcolsep}{2pt}
\small{    \begin{tabular}{lccccccc}
    \toprule
         &
         \textbf{Cora} &  
         \textbf{Citeseer} & 
         \textbf{Pubmed} &
         \textbf{Collab} &
         \textbf{PPA} &
         \textbf{Citation2} 
         &\textbf{DDI} 
         \\
\midrule
          Metric &

          HR@100 &
          HR@100 & 
          HR@100 &
          HR@50 &
          HR@100 &
          MRR 
          &HR@20
         \\ 
         \midrule
         \iffalse
         \textbf{NoTarMaskGAE}& $85.73{\scriptstyle \pm 0.90}$
         & $88.71{\scriptstyle \pm 1.02}$
         & $77.42{\scriptstyle \pm 1.40}$
         & TODO
         &TODO&
         TODO\\
         \fi
         \textbf{CN} & 
         $33.92 {\scriptstyle \pm 0.46}$& 
         $29.79 {\scriptstyle \pm 0.90}$& 
         $23.13 {\scriptstyle \pm 0.15}$&
         $56.44 {\scriptstyle \pm 0.00}$&
         $27.65 {\scriptstyle \pm 0.00}$&
         $51.47 {\scriptstyle \pm 0.00}$
         &$17.73 {\scriptstyle \pm 0.00}$
         \\
\iffalse
         \textbf{FeatCN} &$81.62{\scriptstyle \pm 0.92}$
         & $90.92{\scriptstyle \pm 0.95}$
         & $62.63{\scriptstyle \pm 1.68}$
         & $63.55{\scriptstyle \pm 0.36}$
         &$40.82{\scriptstyle \pm 1.02}$&
         $80.81{\scriptstyle \pm 0.11}$\\
\fi
         \textbf{GAE}& $89.01{\scriptstyle \pm 1.32}$
         & $91.78{\scriptstyle \pm 0.94}$
         & $78.81{\scriptstyle \pm 1.64}$
         & $36.96{\scriptstyle \pm 0.95}$
         & $19.49{\scriptstyle \pm 0.75}$&
         $79.95{\scriptstyle \pm 0.09}$
         &$61.53{\scriptstyle \pm 9.59}$\\
         \textbf{GAE+CN} &$88.61{\scriptstyle \pm 1.31}$
         & $91.75{\scriptstyle \pm 0.98}$
         & $79.04{\scriptstyle \pm 0.83}$
         & $64.47{\scriptstyle \pm 0.14}$
         &$51.83{\scriptstyle \pm 0.58}$&
         $87.81{\scriptstyle \pm 0.06}$
         & $80.71 {\scriptstyle \pm 5.56}$\\
\iffalse    \textbf{GAE+CNC} & $88.37{\scriptstyle \pm 1.22}$
         & $91.66{\scriptstyle \pm 0.95}$
         & $79.22{\scriptstyle \pm 1.04}$
         & $64.72{\scriptstyle \pm 0.40}$
         & $41.62{\scriptstyle \pm 1.68}$&
         $87.53{\scriptstyle \pm 0.04}$
         & $36.28{\scriptstyle \pm 15.98}$\\
\fi
\midrule
         \textbf{NCN2} &
         $88.87{\scriptstyle \pm 1.34}$&
         $91.36{\scriptstyle \pm 1.02}$&
         $80.21{\scriptstyle \pm 0.78}$&
         $65.43{\scriptstyle \pm 0.46}$&
         OOM& 
         OOM&
         OOM\\
         \textbf{NCN-diff} &$89.12{\scriptstyle \pm 1.04}$&
         $91.96{\scriptstyle \pm 1.23}$&
         $80.28{\scriptstyle \pm 0.88}$&
         $64.08{\scriptstyle \pm 0.40}$&
         $57.86{\scriptstyle \pm 1.26}$& 
         $86.68{\scriptstyle \pm 0.16}$&
         $17.67{\scriptstyle \pm 8.70}$
         \\
\midrule
         \textbf{NCN} &
         $89.05{\scriptstyle \pm 0.96}$&
         $91.56{\scriptstyle \pm 1.43}$&
         $79.05{\scriptstyle \pm 1.16}$&
         $64.76{\scriptstyle \pm 0.87}$&
         $61.19{\scriptstyle \pm 0.85}$& 
         $88.09{\scriptstyle \pm 0.06}$&%$88.64{\scriptstyle \pm 0.14}$ & 
         $82.32{\scriptstyle \pm 6.10}$
         \\
         \iffalse
         \textbf{NoTLR}& $85.46{\scriptstyle \pm 1.65}$
         & $88.08{\scriptstyle \pm 1.23}$
         & $76.59{\scriptstyle \pm 1.33}$
         & $64.22{\scriptstyle \pm 0.49}$
         & $60.66{\scriptstyle \pm 0.63}$
         & $88.64{\scriptstyle \pm 0.14}$
         & $66.52{\scriptstyle \pm 11.37}$ 
         \\
         \fi
         \iffalse
         \textbf{GAE}& $89.01{\scriptstyle \pm 1.32}$
         & $91.78{\scriptstyle \pm 0.94}$
         & $78.81{\scriptstyle \pm 1.64}$
         & $36.96{\scriptstyle \pm 0.95}$
         & $19.49{\scriptstyle \pm 0.75}$&
         $78.78{\scriptstyle \pm 0.10}$
         &$61.53{\scriptstyle \pm 9.59}$\\
         \textbf{GAE-TLR}& $85.37{\scriptstyle \pm 0.94}$
         & $89.15{\scriptstyle \pm 1.29}$
         & $77.01{\scriptstyle \pm 1.46}$
         & $39.55{\scriptstyle \pm 1.66}$
         & $19.90{\scriptstyle \pm 0.91}$
         & $79.95{\scriptstyle \pm 0.09}$
         & $66.51{\scriptstyle \pm 6.27}$
         \\
         \fi
         \textbf{NCNC} &
         $89.65{\scriptstyle \pm 1.36}$&
         $93.47{\scriptstyle \pm 0.95}$&
         $81.29{\scriptstyle \pm 0.95}$&
         $66.61{\scriptstyle \pm 0.71}$&
         $61.42{\scriptstyle \pm 0.73}$& 
         $89.12{\scriptstyle \pm 0.40}$ &
         $84.11{\scriptstyle \pm 3.67}$
         \\
         \iffalse
         \textbf{NCNC-2} & $89.14{\scriptstyle \pm 0.84}$
         & $93.14{\scriptstyle \pm 0.96}$
         & $81.41{\scriptstyle \pm 1.07}$
         & $66.80{\scriptstyle \pm 0.43}$
         &$>24h$&
         $>24h$&
         $>24h$
         \\ 
         \fi
         \bottomrule
\end{tabular}}
\vskip -0.15in
\end{table*}

In this section, we extensively evaluate the performance of both NCN and NCNC. Detailed experimental settings are included in Appendix~\ref{app::experimentsetting}. 

We use seven popular real-world link prediction benchmarks. Among these, three are Planetoid citation networks: Cora, Citeseer, and Pubmed~\citep{Cora}. Others are from Open Graph Benchmark~\citep{OGB}: ogbl-collab, ogbl-ppa, ogbl-citation2, and ogbl-ddi. Their statistics and splits are shown in Appendix~\ref{app:data}. 

\subsection{Evaluation on Real-World Datasets}

In our evaluation on real-world datasets, we employ a range of baseline methods, encompassing traditional heuristics like CN~\citep{CommonNeighbor}, RA~\citep{RA}, and AA~\citep{AA}, as well as GAE models, such as GCN~\citep{GCN} and SAGE~\citep{GraphSage}. Additionally, we consider SF-then-MPNN models, including SEAL~\citep{SEAL} and NBFNet~\citep{NBFNet}, as well as SF-and-MPNN models like Neo-GNN~\citep{Neo-GNN} and BUDDY~\citep{Gsketch}. The baseline results are sourced from~\citep{Gsketch}. Our models consist of NCN and NCNC. Their architectures are detailed in Appendix~\ref{app:arch}.

The experimental results are presented in Table~\ref{tab:main_results}. NCN surpasses all baselines on 5/7 datasets and exhibits an average score improvement of 5\% compared to BUDDY, the most competitive baseline. Even on the remaining two datasets, NCN outperforms all baselines except BUDDY. These impressive results underscore the outstanding expressivity of our MPNN-then-SF architecture. Furthermore, NCNC enhances performance by an additional 2\%, emerging as the top-performing method on all datasets. Notably, on ogbl-ppa, NCNC achieves an HR@100 score of 61.42\%, surpassing the strongest baseline BUDDY by a substantial margin of over 10\%. It's worth mentioning that our models outperform node embedding techniques~\citep{Deepwalk,Node2Vec,Line} and other GNNs lacking pairwise features~\citep{PLNLP,GIDN} significantly (see Appendix~\ref{app:plnlp}).

% Figure 1, line width match text
% textit
% interval of formula

\subsection{Scalability}\label{sec:scalability}
\begin{wrapfigure}{r}{0.69\textwidth}
  \begin{center}\vskip -0.1in
    \includegraphics[width=0.69\textwidth]{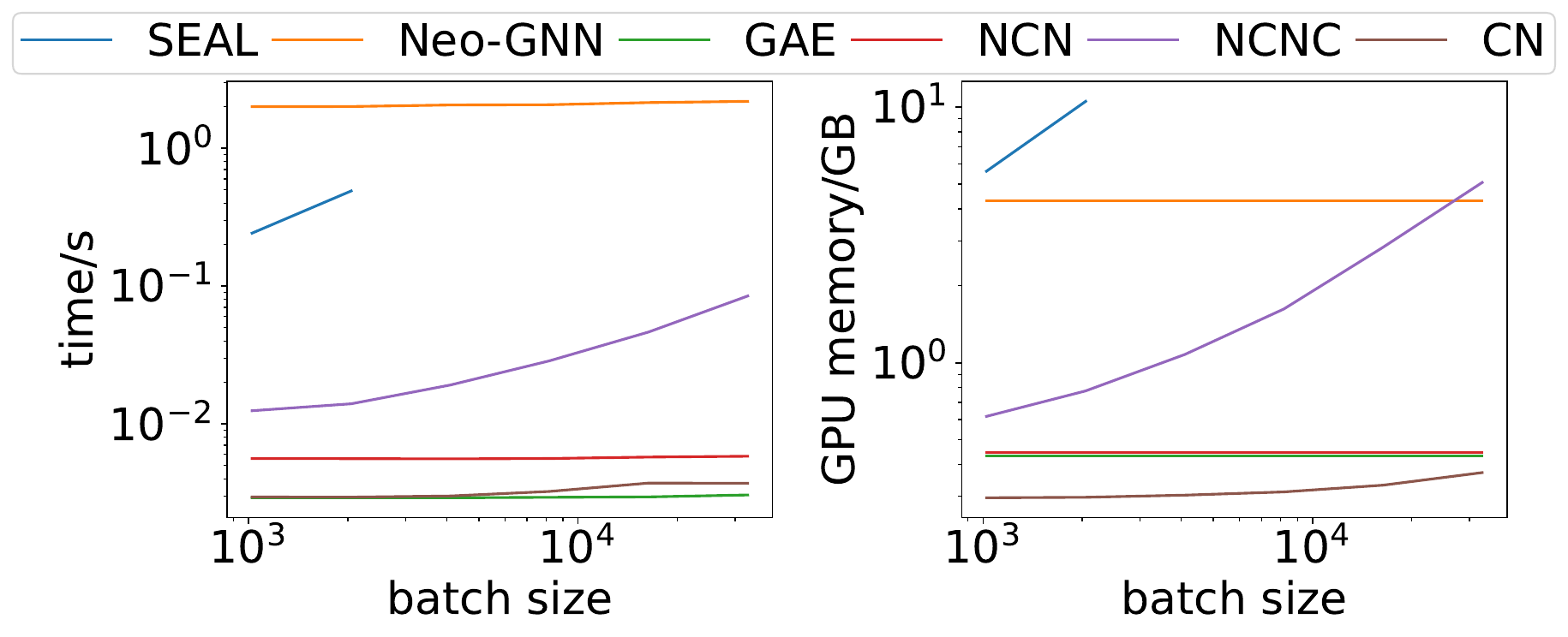}
  \end{center}
    \vskip -0.15in
    \caption{Inference time and GPU memory on ogbl-collab. The process we measure includes preprocessing and predicting one batch of test links. As shown in Appendix~\ref{app:complexity}, relation between time $y$ and batch size $t$ is $y=B+Ct$, where $B,C$ are model-specific constants. SEAL has out-of-memory problem and only uses small batch sizes.}\label{fig:time}
\vskip -0.15in
\end{wrapfigure}

%As shown in Appendix~\ref{app:complexity}, the time complexity of link prediction models can be expressed as $O(A+Bt)$, where $A,B$ are model-specific parameters and $t$ is the number of edges.
We compare the inference time and GPU memory on ogbl-collab in Figure~\ref{fig:time}. NCN and NCNC have a \textbf{similar computation overhead to GAE}, as they both need to run MPNN only once. 
In contrast, SEAL, which reruns MPNN for each target link, takes $86$ times more time compared with NCN with a small batch size $2048$, and the disadvantage will be more significant with a larger batch size. Surprisingly, BUDDY and Neo-GNN are slower than NCN. The reason is that it uses pairwise features depending on high order neighbors that are much more time-consuming than common neighbor. NCN and NCNC also achieve low GPU memory consumption. We also conduct scalability comparisons on other datasets and observe the same results (see Appendix~\ref{app:time}).

\subsection{Ablation Analysis}\label{sec::abl}
To assess the effectiveness of the NCNC design, we conducted a comprehensive ablation analysis, as presented in Table~\ref{tab:abl}.

Starting with GAE, which relies solely on node representations, we introduced GAE+CN, which incorporates Common Neighbor (CN) as pairwise features. Remarkably, GAE+CN outperforms GAE by 70\% on Open Graph Benchmark (OGB) datasets, illustrating the importance of structural features. Furthermore, NCN exhibits a 5.5\% score increase over GAE+CN, highlighting the advantages of the MPNN-then-SF architecture over the MPNN-and-SF architecture.

%\jr{In the following three paragraphs, the narrative is in past tense, inconsistent with the remaining parts of the paper.}
We also explore variants of NCN, namely NCN-diff and NCN2. In NCN-diff, we include neighborhood difference information by summing the representations of nodes in $N(i, A)-N(j, A)$ and $N(j, A)-N(i, A)$, while NCN2 incorporates high-order neighborhood overlap using $N(i, A^2)\cap N(j, A)$ and $N(i, A)\cap N(j, A^2)$. Notably, NCN, NCN-diff, and NCN2 exhibit similar performances across most datasets, suggesting that first-order neighborhood overlap might be sufficient. However, NCN-diff achieves a lower score on the DDI dataset, possibly because the high node degree in DDI introduces noisy and uninformative neighborhood difference information.

\section{Conclusion}
In this work, we introduce Neural Common Neighbor (NCN), a scalable and robust model for link prediction that harnesses the power of learnable pairwise features. Additionally, we address the challenge of graph incompleteness by identifying and visualizing common neighbor loss and distribution shifts stemming from this issue. To mitigate these problems, we introduce the Common Neighbor Completion (CNC) technique. Combining CNC with NCN, our final model, Neural Common Neighbor with Completion (NCNC), outperforms state-of-the-art baselines across various datasets in terms of both speed and prediction performance.

\section{Limitations}
Though we propose MPNN-then-SF framework, we do not exhaust the design space and only propose one implementation, NCN, and its variants NCN2 and NCN-diff in ablation study. Moreover, while we only analyze the impact of incompleteness on common neighbor structures, graph incompleteness can also affect other structural features. Additionally, the proposed completion method has the potential to be generalized to address other structural features. Our future research will explore the design space of MPNN-then-SF and the broader implications of incompleteness on various structural features. 

\section{Reproducibility Statement}
Our code is available at \url{https://github.com/GraphPKU/NeuralCommonNeighbor}. Proofs of all theorems in the maintext are in Appendix~\ref{app::proof}.

\section*{Acknowledgement}
This work is partially supported by the National Key R\&D Program of China (2022ZD0160303), the National Natural Science Foundation of China (62276003), and Alibaba Innovative Research Program.

\bibliography{iclr2024_conference}

\begin{thebibliography}{31}
\providecommand{\natexlab}[1]{#1}
\providecommand{\url}[1]{\texttt{#1}}
\expandafter\ifx\csname urlstyle\endcsname\relax
  \providecommand{\doi}[1]{doi: #1}\else
  \providecommand{\doi}{doi: \begingroup \urlstyle{rm}\Url}\fi

\bibitem[Adamic \& Adar(2003)Adamic and Adar]{AA}
Lada~A Adamic and Eytan Adar.
\newblock Friends and neighbors on the web.
\newblock \emph{Social networks}, 25\penalty0 (3):\penalty0 211--230, 2003.

\bibitem[Akiba et~al.(2019)Akiba, Sano, Yanase, Ohta, and Koyama]{optuna}
Takuya Akiba, Shotaro Sano, Toshihiko Yanase, Takeru Ohta, and Masanori Koyama.
\newblock Optuna: A next-generation hyperparameter optimization framework.
\newblock In \emph{KDD}, pp.\  2623--2631, 2019.

\bibitem[Barab{\'a}si \& Albert(1999)Barab{\'a}si and Albert]{CommonNeighbor}
Albert-L{\'a}szl{\'o} Barab{\'a}si and R{\'e}ka Albert.
\newblock Emergence of scaling in random networks.
\newblock \emph{science}, 286\penalty0 (5439):\penalty0 509--512, 1999.

\bibitem[Chamberlain et~al.(2023)Chamberlain, Shirobokov, Rossi, Frasca, Markovich, Hammerla, Bronstein, and Hansmire]{Gsketch}
Benjamin~Paul Chamberlain, Sergey Shirobokov, Emanuele Rossi, Fabrizio Frasca, Thomas Markovich, Nils Hammerla, Michael~M. Bronstein, and Max Hansmire.
\newblock Graph neural networks for link prediction with subgraph sketching.
\newblock \emph{ICLR}, 2023.

\bibitem[Dong et~al.(2022)Dong, Tian, Guo, Yang, and Chawla]{FakeEdge}
Kaiwen Dong, Yijun Tian, Zhichun Guo, Yang Yang, and Nitesh~V. Chawla.
\newblock Fakeedge: Alleviate dataset shift in link prediction.
\newblock In \emph{LoG}, volume 198, pp.\ ~56. {PMLR}, 2022.

\bibitem[Fey \& Lenssen(2019)Fey and Lenssen]{pyg}
Matthias Fey and Jan~Eric Lenssen.
\newblock Fast graph representation learning with pytorch geometric.
\newblock \emph{arXiv preprint arXiv:1903.02428}, 2019.

\bibitem[Gilmer et~al.(2017)Gilmer, Schoenholz, Riley, Vinyals, and Dahl]{MPNN}
Justin Gilmer, Samuel~S Schoenholz, Patrick~F Riley, Oriol Vinyals, and George~E Dahl.
\newblock Neural message passing for quantum chemistry.
\newblock In \emph{ICML}, pp.\  1263--1272, 2017.

\bibitem[Grover \& Leskovec(2016)Grover and Leskovec]{Node2Vec}
Aditya Grover and Jure Leskovec.
\newblock node2vec: Scalable feature learning for networks.
\newblock In \emph{KDD}, pp.\  855--864. {ACM}, 2016.

\bibitem[Hamilton et~al.(2017)Hamilton, Ying, and Leskovec]{GraphSage}
William~L Hamilton, Rex Ying, and Jure Leskovec.
\newblock Inductive representation learning on large graphs.
\newblock \emph{NeurIPS}, pp.\  1025--1035, 2017.

\bibitem[Hu et~al.(2020)Hu, Fey, Zitnik, Dong, Ren, Liu, Catasta, and Leskovec]{OGB}
Weihua Hu, Matthias Fey, Marinka Zitnik, Yuxiao Dong, Hongyu Ren, Bowen Liu, Michele Catasta, and Jure Leskovec.
\newblock Open graph benchmark: Datasets for machine learning on graphs.
\newblock In \emph{NeurIPS}, 2020.

\bibitem[Kipf \& Welling(2016)Kipf and Welling]{GAE}
Thomas~N. Kipf and Max Welling.
\newblock Variational graph auto-encoders.
\newblock \emph{CoRR}, abs/1611.07308, 2016.

\bibitem[Kipf \& Welling(2017)Kipf and Welling]{GCN}
Thomas~N. Kipf and Max Welling.
\newblock Semi-supervised classification with graph convolutional networks.
\newblock In \emph{ICLR}, 2017.

\bibitem[Liben-Nowell \& Kleinberg(2003)Liben-Nowell and Kleinberg]{liben2003link}
David Liben-Nowell and Jon Kleinberg.
\newblock The link prediction problem for social networks.
\newblock In \emph{International conference on Information and knowledge management}, pp.\  556--559, 2003.

\bibitem[Paszke et~al.(2019)Paszke, Gross, Massa, Lerer, Bradbury, Chanan, Killeen, Lin, Gimelshein, Antiga, et~al.]{pytorch}
Adam Paszke, Sam Gross, Francisco Massa, Adam Lerer, James Bradbury, Gregory Chanan, Trevor Killeen, Zeming Lin, Natalia Gimelshein, Luca Antiga, et~al.
\newblock Pytorch: An imperative style, high-performance deep learning library.
\newblock \emph{NeurIPS}, 32, 2019.

\bibitem[Perozzi et~al.(2014)Perozzi, Al{-}Rfou, and Skiena]{Deepwalk}
Bryan Perozzi, Rami Al{-}Rfou, and Steven Skiena.
\newblock Deepwalk: online learning of social representations.
\newblock In \emph{KDD}, pp.\  701--710. {ACM}, 2014.

\bibitem[Souri et~al.(2022)Souri, Laddach, Karagiannis, Papageorgiou, and Tsoka]{DrugInteraction}
E.~Amiri Souri, Roman Laddach, S.~N. Karagiannis, Lazaros~G. Papageorgiou, and Sophia Tsoka.
\newblock Novel drug-target interactions via link prediction and network embedding.
\newblock \emph{{BMC} Bioinform.}, 23\penalty0 (1):\penalty0 121, 2022.

\bibitem[Tang et~al.(2015)Tang, Qu, Wang, Zhang, Yan, and Mei]{Line}
Jian Tang, Meng Qu, Mingzhe Wang, Ming Zhang, Jun Yan, and Qiaozhu Mei.
\newblock {LINE:} large-scale information network embedding.
\newblock In Aldo Gangemi, Stefano Leonardi, and Alessandro Panconesi (eds.), \emph{WWW}, pp.\  1067--1077, 2015.

\bibitem[Velickovic et~al.(2018)Velickovic, Cucurull, Casanova, Romero, Li{\`{o}}, and Bengio]{GAT}
Petar Velickovic, Guillem Cucurull, Arantxa Casanova, Adriana Romero, Pietro Li{\`{o}}, and Yoshua Bengio.
\newblock Graph attention networks.
\newblock In \emph{ICLR}, 2018.

\bibitem[Wang et~al.(2021)Wang, Zhou, Hong, Zou, Su, and Chen]{PLNLP}
Zhitao Wang, Yong Zhou, Litao Hong, Yuanhang Zou, Hanjing Su, and Shouzhi Chen.
\newblock Pairwise learning for neural link prediction.
\newblock \emph{CoRR}, abs/2112.02936, 2021.

\bibitem[Wang et~al.(2022)Wang, Guo, Zhao, Zhang, Yu, Liao, Jin, Wang, and Yu]{GIDN}
Zixiao Wang, Yuluo Guo, Jin Zhao, Yu~Zhang, Hui Yu, Xiaofei Liao, Hai Jin, Biao Wang, and Ting Yu.
\newblock {GIDN:} {A} lightweight graph inception diffusion network for high-efficient link prediction.
\newblock \emph{CoRR}, abs/2210.01301, 2022.

\bibitem[Xu et~al.(2019)Xu, Hu, Leskovec, and Jegelka]{HowPowerfulAreGNNs}
Keyulu Xu, Weihua Hu, Jure Leskovec, and Stefanie Jegelka.
\newblock How powerful are graph neural networks?
\newblock In \emph{ICLR}, 2019.

\bibitem[Yang et~al.(2022)Yang, Lin, and Zhang]{OpenworldKG}
Haotong Yang, Zhouchen Lin, and Muhan Zhang.
\newblock Rethinking knowledge graph evaluation under the open-world assumption.
\newblock pp.\  13683--13694, 2022.

\bibitem[Yang et~al.(2016)Yang, Cohen, and Salakhutdinov]{Cora}
Zhilin Yang, William~W. Cohen, and Ruslan Salakhutdinov.
\newblock Revisiting semi-supervised learning with graph embeddings.
\newblock In \emph{Proceedings of the 33nd International Conference on Machine Learning}, volume~48, pp.\  40--48, 2016.

\bibitem[Yun et~al.(2021)Yun, Kim, Lee, Kang, and Kim]{Neo-GNN}
Seongjun Yun, Seoyoon Kim, Junhyun Lee, Jaewoo Kang, and Hyunwoo~J. Kim.
\newblock Neo-gnns: Neighborhood overlap-aware graph neural networks for link prediction.
\newblock In \emph{NeurIPS}, pp.\  13683--13694, 2021.

\bibitem[Zhang \& Chen(2018)Zhang and Chen]{SEAL}
Muhan Zhang and Yixin Chen.
\newblock Link prediction based on graph neural networks.
\newblock \emph{NeurIPS}, 31:\penalty0 5165--5175, 2018.

\bibitem[Zhang \& Chen(2020)Zhang and Chen]{Zhang2020}
Muhan Zhang and Yixin Chen.
\newblock Inductive matrix completion based on graph neural networks.
\newblock In \emph{ICLR}, 2020.

\bibitem[Zhang et~al.(2021)Zhang, Li, Xia, Wang, and Jin]{zhang2021labeling}
Muhan Zhang, Pan Li, Yinglong Xia, Kai Wang, and Long Jin.
\newblock Labeling trick: A theory of using graph neural networks for multi-node representation learning.
\newblock \emph{NeurIPS}, 34:\penalty0 9061--9073, 2021.

\bibitem[Zhao et~al.(2023)Zhao, Wen, Ju, Zhang, and Ye]{Gcomplete1}
Jianan Zhao, Qianlong Wen, Mingxuan Ju, Chuxu Zhang, and Yanfang Ye.
\newblock Self-supervised graph structure refinement for graph neural networks.
\newblock In \emph{WSDM}, pp.\  159--167, 2023.

\bibitem[Zhao et~al.(2021)Zhao, Liu, Neves, Woodford, Jiang, and Shah]{Gcomplete2}
Tong Zhao, Yozen Liu, Leonardo Neves, Oliver~J. Woodford, Meng Jiang, and Neil Shah.
\newblock Data augmentation for graph neural networks.
\newblock In \emph{AAAI}, pp.\  11015--11023, 2021.

\bibitem[Zhou et~al.(2009)Zhou, L{\"u}, and Zhang]{RA}
Tao Zhou, Linyuan L{\"u}, and Yi-Cheng Zhang.
\newblock Predicting missing links via local information.
\newblock \emph{The European Physical Journal B}, 71\penalty0 (4):\penalty0 623--630, 2009.

\bibitem[Zhu et~al.(2021)Zhu, Zhang, Xhonneux, and Tang]{NBFNet}
Zhaocheng Zhu, Zuobai Zhang, Louis{-}Pascal A.~C. Xhonneux, and Jian Tang.
\newblock Neural bellman-ford networks: {A} general graph neural network framework for link prediction.
\newblock In \emph{NeurIPS}, pp.\  29476--29490, 2021.

\end{thebibliography}
\bibliographystyle{iclr2024_conference}
\appendix

\newpage
\appendix
\onecolumn

\section{Proof}\label{app::proof}

\subsection{Derivation of Equation~\ref{equ:GenPairwise3}}\label{app::proof::equ}
The MPNN-then-SF architecture is as follows,
\begin{equation}
    \text{Pool}(\msl \text{MPNN}(u, A, X)|u\in S \msr)
\end{equation}
Let $S_{ab}$ be the following set,
\begin{equation}
    (N_{l_1}^{l_2}(i)\oplus N_{l_1'}^{l_2'}(j)) \cap \{u\in V|A^{l_2}_{iu}=a\} \cap \{u \in V|A^{l_2'}_{uj}=b\}.
\end{equation}
Then, for $S_{ab}$, we set the pooling function to sum and multiplied with $g(a)g(b)$, where $g$ is a function with high-order adjacency edge weight as input. Then the MPNN-then-SF architecture can express,
\begin{equation}
    \sum_{u\in S_{ab}}g(a)g(b)\text{MPNN}(u, A, X)
\end{equation}
Simply sums the feature of all $S_{ab}$ leads to,
\begin{equation}
    \sum_{u\in N_{l_1}^{l_2}(i)\oplus N_{l_1'}^{l_2'}(j)} g(A_{iu}^{l_2})g(A_{ju}^{l_2'}) \text{MPNN}(u, A, X).
\end{equation}

\subsection{Proof of Theorem~\ref{thm::expressivity} and ~\ref{thm::expressivity2}}

Here, we present the theoretical proof of MPNN-then-SF's higher expressivity. We say algorithm A is strictly more expressive than algorithm B when A can differentiate all pairs of links that B can differentiate, while there exists a pair of links that A can distinguish while B cannot. We first prove the more expressive results by simulating other models with SF-then-MPNN and NCN then prove the strictness by constructing an example.

\begin{lemma}
Equation~\ref{equ:GenPairwise2} and NCN are more expressive than Graph Autoencoder (GAE)
\end{lemma}
\begin{proof}
Graph Autoencoder's prediction for link (i, j) is $\langle \text{MPNN}(i, A, X), \text{MPNN}(j, A, X)\rangle$. So directly sum Equation~\ref{equ:GenPairwise2} leads to GAE. Equation~\ref{equ:GenPairwise2} is a part of NCN, so NCN can also express GAE.
\end{proof}

\begin{lemma}
    NCN is more expressive than CN,RA, and AA. Combination of Equation~\ref{equ:GenPairwise2} and Equation~\ref{equ:GenPairwise3} is more expressive than CN,RA,AA, BUDDY and Neo-GNN
\end{lemma}
\begin{proof}
As MPNN can learn arbitrary functions of node degrees, NCN can express Equation~\ref{equ:heuristics}, and Equation~\ref{equ:GenPairwise3} can express the general form of structure feature~\ref{equ:framework}. 
\end{proof}

Furthermore, we construct an example in Figure~\ref{fig::NCNexpexp}. In that graph, $v_2$ and $v_3$ are symmetric and thus have the same MPNN representation, so GAE cannot distinguish $(v_1,v_2)$ and $(v_1,v_3)$. Moreover, $(v_1,v_2)$ and $(v_1,v_3)$ are symmetric if the node feature is ignored, so CN, RA, AA, Neo-GNN, BUDDY cannot distinguish them. However, $(v_1,v_2)$ have a common neighbor with feature $2$, and $(v_1,v_3)$ have a common neighbor with feature $1$, so NCN can distinguish them. 

\section{Dataset Statistics}\label{app:data}
The statistics of each dataset are shown in Table~\ref{tab:datasets}.

\begin{table*}[t]
    \centering
    \caption{Statistics of dataset.}\label{tab:datasets}
    \vskip 0.15in
    \begin{tabular}{l ccccccc}
    \toprule 
         &
         \textbf{Cora} &  
         \textbf{Citeseer} & 
         \textbf{Pubmed} &
         \textbf{Collab} &
         \textbf{PPA} &
         \textbf{DDI} &
         \textbf{Citation2}
        \\
\midrule
         
                  \#Nodes &

         2,708 & 
         3,327 &
         18,717 &
         235,868 &
         576,289 &
         4,267 &
         2,927,963
          \\
         
         \#Edges &
         5,278 & 
         4,676 &
         44,327 &
         1,285,465 &
         30,326,273 &
         1,334,889 &
         30,561,187
          \\

         splits &

         random &
         random & 
         random &
         fixed &
         fixed &
         fixed &
         fixed \\
          
         average degree &
         3.9 &
         2.74 & 
         4.5 &
         5.45 &
         52.62 &
         312.84 &
         10.44
        \\
         
         \bottomrule
\end{tabular}

\label{tab:subgraph properties}
\end{table*}
Random splits use $70\%/10\%/20\%$ edges for training/validation/test set respectively. Different from others, the collab dataset allows using validation edges as input on test set. 

\section{Model Architecture}\label{app:arch}

This section concludes our methods in Section~\ref{sec:NCN} and Section~\ref{sec::intervene}.

\begin{figure}
    \centering
    \includegraphics[width=\textwidth]{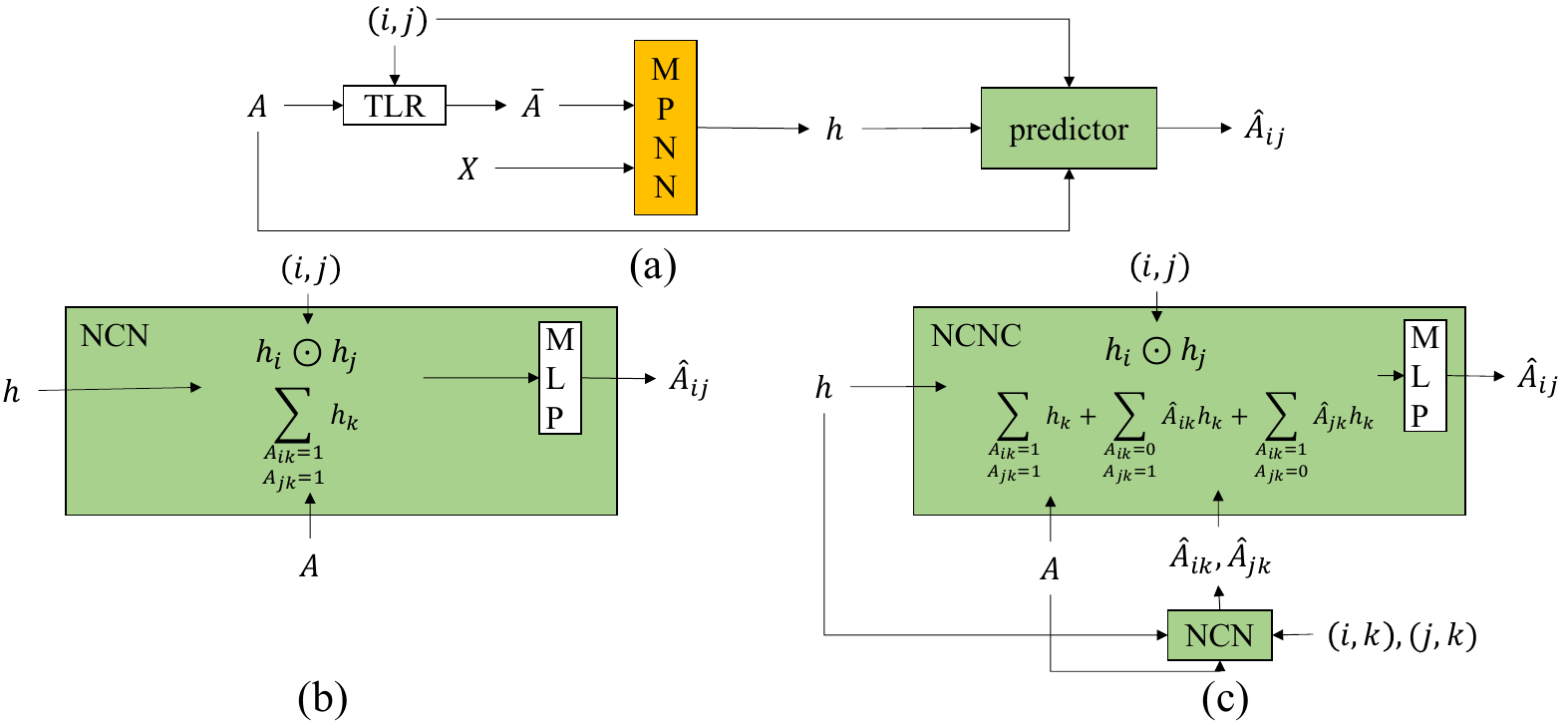}
    \caption{Architecture of our models. (a) The overall architecture. Given node feature $X$, adjacency matrix $A$ and target link $(i,j)$, models first set $A_{ij}=0$ ( Target link removal, TLR). Then, $\bar A, X$ are fed to a vanilla MPNN for node representations $h$. With $(i, j)$, $h$, and $A$ as input, the predictor produces $\hat A_{ij}$, the probability that edge $(i,j)$ exists. (b) The NCN predictor. It uses node representations of target nodes $i, j$ and their common neighbors to produce edge representations. Then, it feeds the edge representation to an MLP to produce the final prediction. (c) The NCNC predictor. It first uses NCN to predict unobserved links $\hat A_{ik},\hat A_{jk}$, which is then used to complete unobserved common neighbors. }
    \label{fig:arch}
\end{figure}

Given an input graph $A$, a node feature matrix $X$, and target links $\{(i_1, j_1), (i_2, j_2),..., (i_t, j_t)\}$, our models consist of three steps: target link removal, MPNN, and predictor. NCN and NCNC only differ in the last step. The model architecture is visualized in Figure~\ref{fig:arch}

\paragraph{Target link removal.} We make no changes to the input graph in the validation and test set where the target links are unobserved. In the training set, we remove target links from $A$. Let $\bar A$ denote the processed graph. This method is detailed in Section~\ref{sec::intervene}.

\paragraph{MPNN.} We use MPNN to produce node representations $h$. For each node $i$,
\begin{equation}
    h_i=\text{MPNN}(i, \bar A, X).
\end{equation}
For all target links, MPNN needs to run only once.

\paragraph{Predictor.}
Predictors use the node representations and graph structure to produce link prediction. Link representations of NCN are as follows,
\begin{equation}\label{equ:ncn}
    z_{ij}=(h_i\odot h_j||\sum_{u\in N(i, \bar A)\cap\atop N(j, \bar A)} h_u),
\end{equation}
where $||$ means concatenation, $z_{ij}$ is the  representation of link $(i, j)$. $z_{ij}$ composed of two components: two nodes' presentation $h_i\odot h_j$ and representations of nodes within the common neighbor set. The former component is often used in link prediction models~\citep{GAE, Neo-GNN, Gsketch}, while we propose the latter one for the first time. Link representations are then used to produce link existence probability.
\begin{equation}
    \hat A_{ij}=\text{sigmoid}(\text{MLP}(z_{ij})),
\end{equation}
where $\hat A_{ij}$ is the probability that link $(i, j)$ exists. 

NCNC has a similar form. The only difference is that $\sum_{u\in N(i, \bar A)\cap N(j, \bar A)} h_u$ in Equation~(\ref{equ:ncn}) is replaced with the follow form:
\begin{equation}
    \sum_{u\in N(i, \bar A)\cap 
    \atop N(j, \bar A)} h_u + 
    \sum_{u\in N(j, \bar A)-
    \atop N(i, \bar A)} \hat A_{iu} h_u +
    \sum_{u\in N(i, \bar A)- 
    \atop N(j, \bar A)} \hat A_{ju} h_u.
\end{equation}
where $\hat A_{ab}$ is the link existence probability produced by NCNC. 

\section{Experimental Settings}\label{app::experimentsetting}

{\bf Computing infrastructure.}~~We leverage Pytorch Geometric~\citep{pyg} and Pytorch~\citep{pytorch} for model development. All experiments are conducted on an Nvidia 4090 GPU on a Linux server. 

{\bf Baselines.}~~We directly use the results reported in \citep{Gsketch}. 

{\bf Model hyperparameter.}~~We use optuna~\citep{optuna} to perform random searches. Hyperparameters were selected to maximize validation score. The best hyperparameters selected for each model can be found in our code. 

{\bf Training process.} We utilize Adam optimizer to optimize models and set an epoch upper bound $100$. All results of our models are provided from runs with 10 random seeds.

{\bf Computation cost} The total time of each main experiment is shown in Table~\ref{tab:cost}. Reproducing all main results takes 280 GPU hours.

\begin{table*}[t]
    \centering
    \caption{Total time(s) needed in one run}\label{tab:cost}
\vskip 0.15in
\setlength{\tabcolsep}{2pt}
\small{
    \begin{tabular}{lccccccc}
    \toprule
         &
         \textbf{Cora} &  
         \textbf{Citeseer} & 
         \textbf{Pubmed} &
         \textbf{Collab} &
         \textbf{PPA} &
         \textbf{Citation2} 
         &\textbf{DDI} 
         \\
         \midrule
         \textbf{NCN} &
         $8$&
         $16$&
         $28$&
         $320$&
         $9375$& 
         $7123$&
         $546$
         \\
        \textbf{NCNC} &
         $15$&
         $27$&
         $54$&
         $730$&
         $77385$& 
         $5170$&
         $1785$
         \\ \bottomrule
\end{tabular}
}
\end{table*}

\section{Time and Space Complexity}\label{app:complexity}
\begin{table}[t]
    \centering
    \caption{Scalability comparison. $h,h, h''$: the complexity of hash function in BUDDY, where are all $\ge d$. $F$: the dimension of node representations. When predicting the $t$ target links, time and space complexity of existing models can be expressed as $O(B+Ct)$ and $O(D+Et)$ respectively.}\label{tab:complexity}
    %\vskip 0.15in
\small{
    \begin{tabular}{llcccc}
    \toprule
    Architecture&Method& B & C& D & E\\
    \midrule
    MPNN only &GAE &$ndF+nF^2$ & $F^2$& $nF$ & $F$ \\ 
    \midrule    
    MPNN-and-SF&Neo-GNN & $ndF+nF^2+nd^l$ &$d^{l}+F^2$& $nF+nd^l$ & $d^l+F$\\
    &BUDDY &$ndF+nh$ & $h'+F^2$&$nF+nh''$ & $F+h'$ \\
    \midrule
    SF-then-MPNN&SEAL &$0$ & $d^{l'+1}F+d^{l'}F^2$& $0$& $d^{l'+1}F$ \\
    \midrule
    MPNN-then-SF&NCN &$ndF+nF^2$ & $dF+F^2$& $nF$ & $dF$\\ 
    &NCNC &$ndF+nF^2$ & $d^2F+dF^2$ &$nF$ & $d^2F$\\
    %NCNC-$K$ & $ndF+nF^2$ & $d^{K+1}F+d^KF^2$& $nF$& $d^K F$\\
    \bottomrule
\end{tabular}
}
\vskip -0.1in
\end{table}
\begin{figure}[h]
% \vskip 0.2in
    \centering
    \subfloat[Cora]{\includegraphics[width=0.5\textwidth]{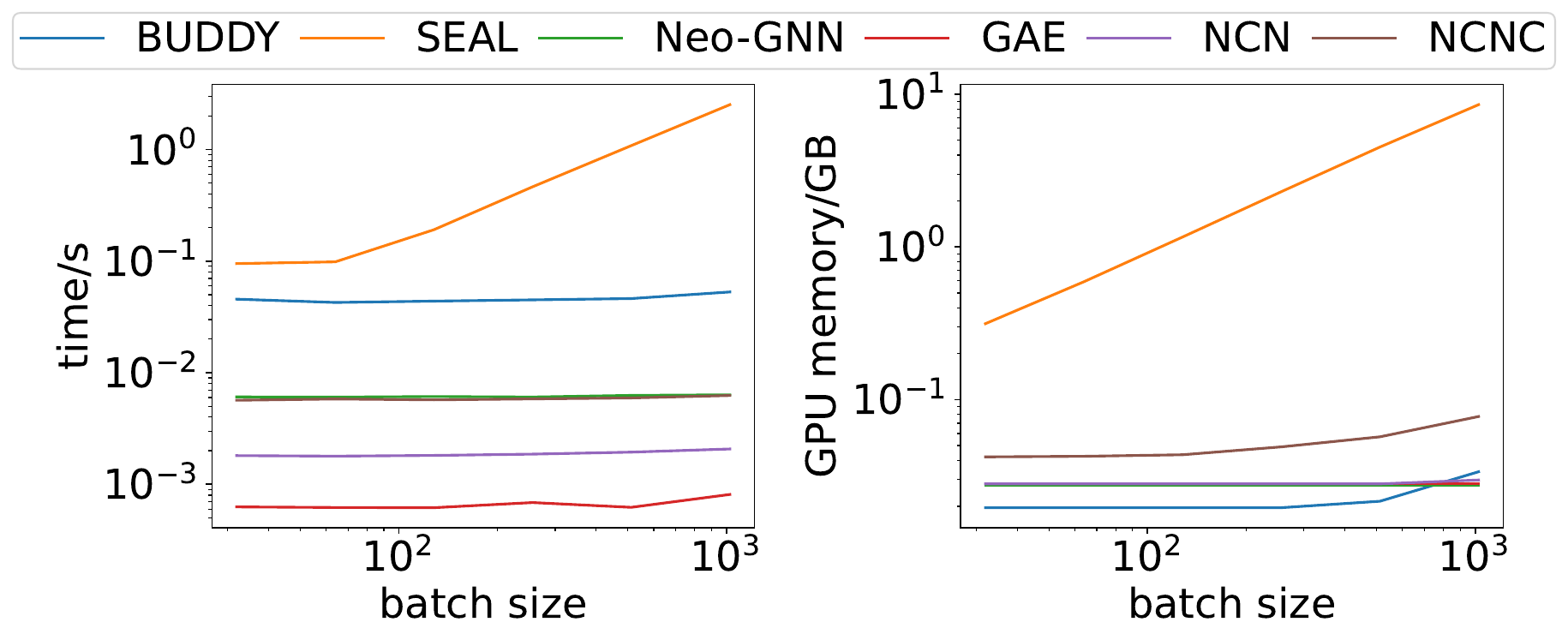}}
   \subfloat[Citeseer]{ \includegraphics[width=0.5\textwidth]{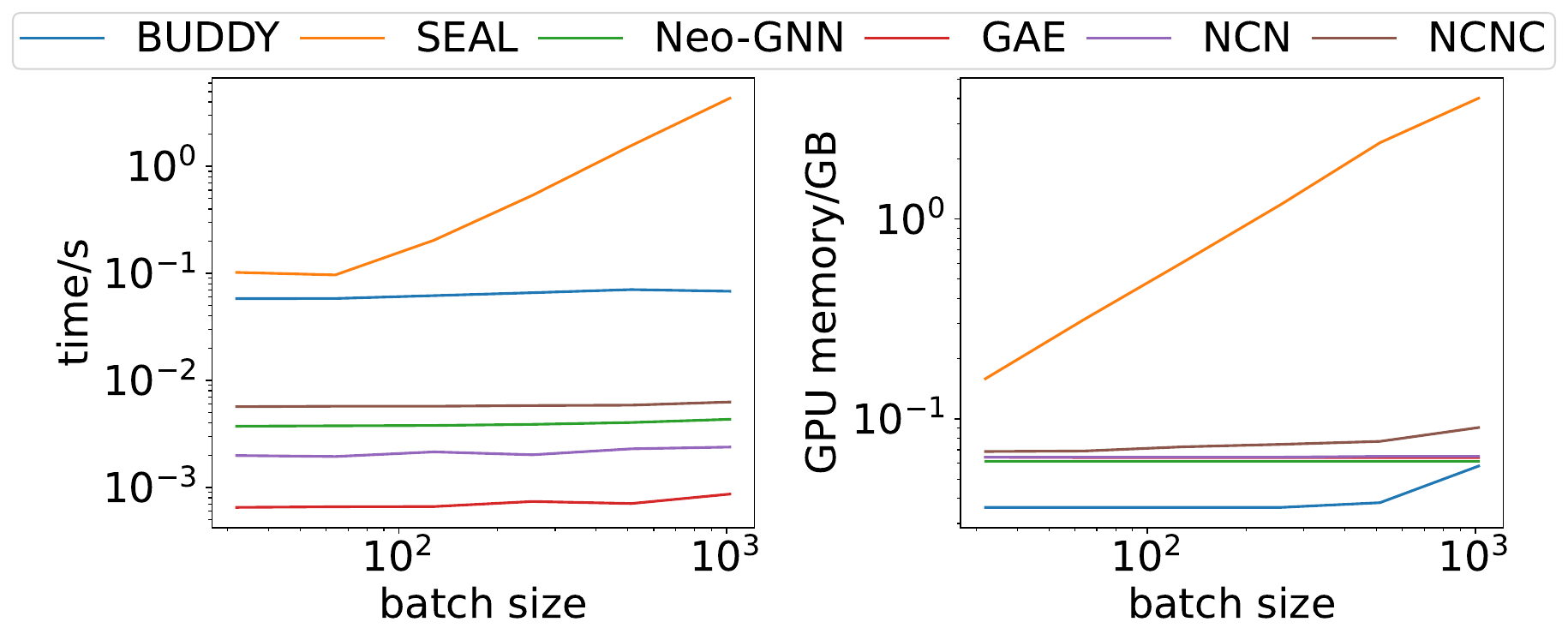}}\quad
   \subfloat[Pubmed]{ \includegraphics[width=0.5\textwidth]{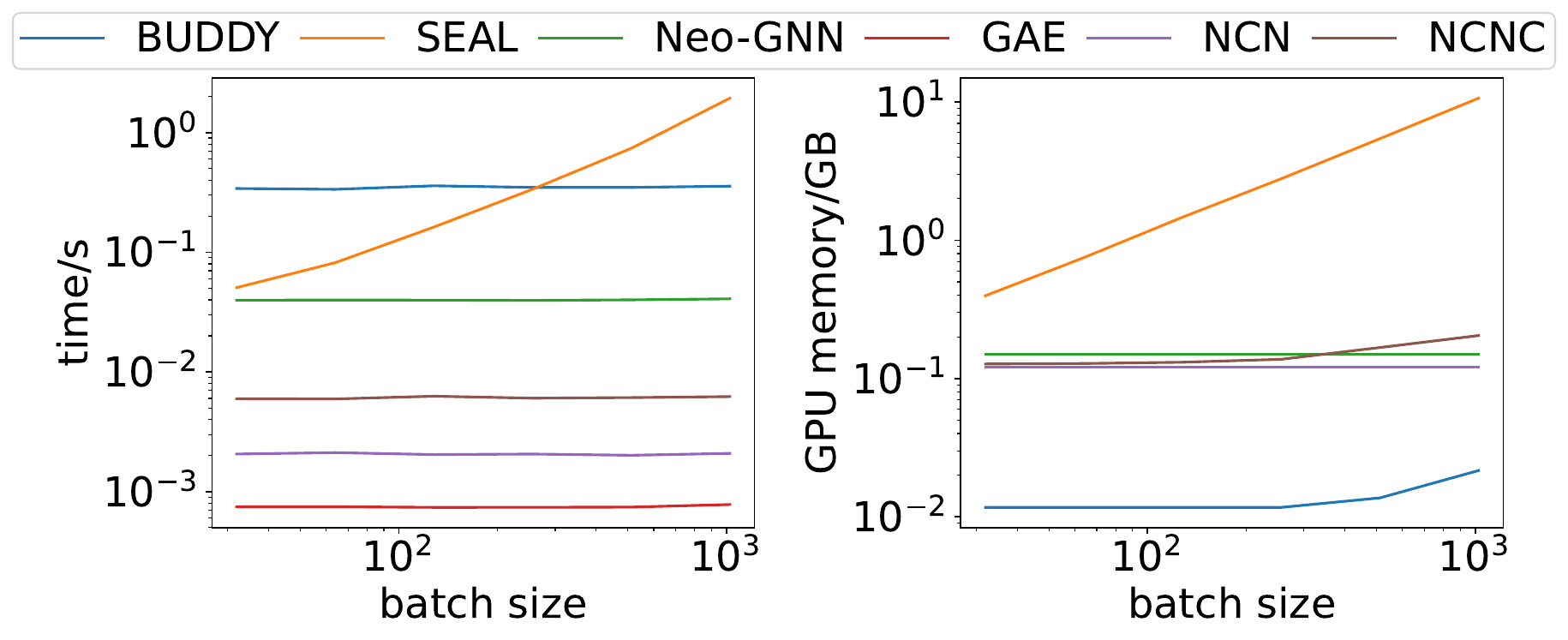}}
   \subfloat[Ogbl-ppa]{ \includegraphics[width=0.5\textwidth]{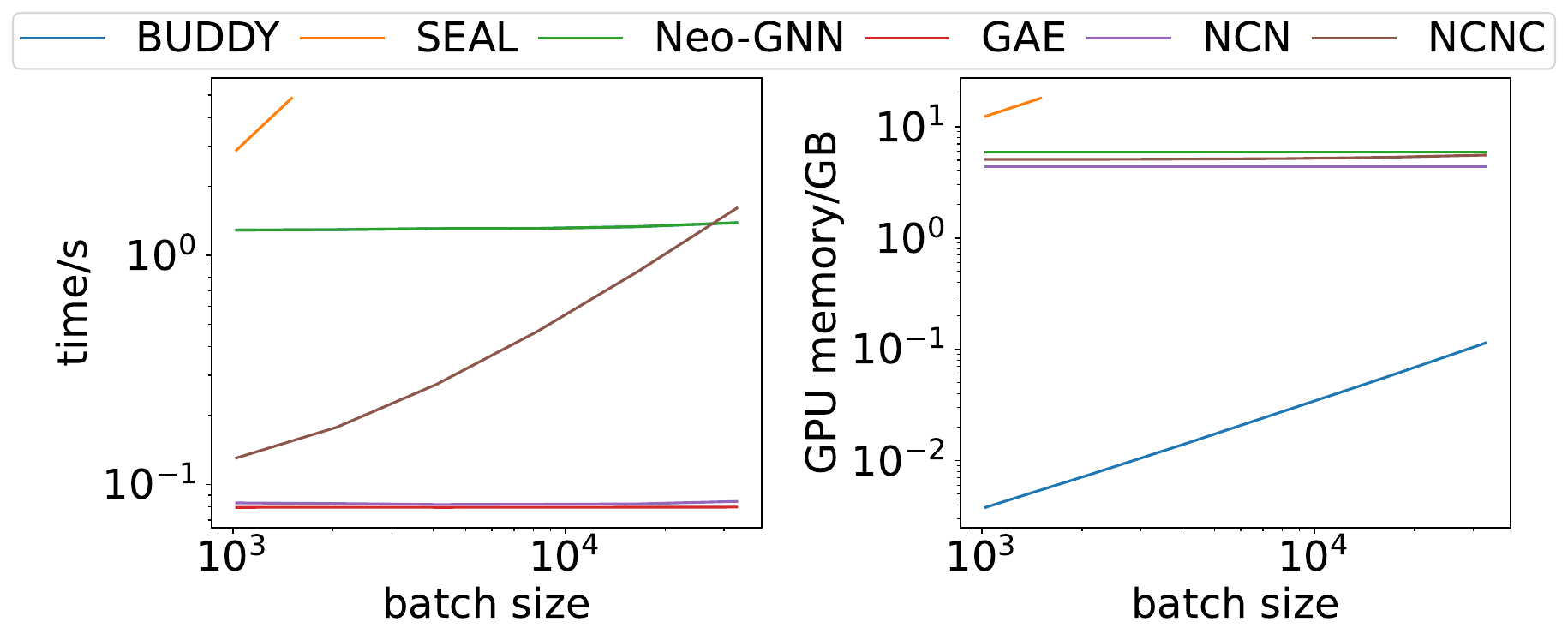}}\quad
    \subfloat[Ogbl-ddi]{ \includegraphics[width=0.5\textwidth]{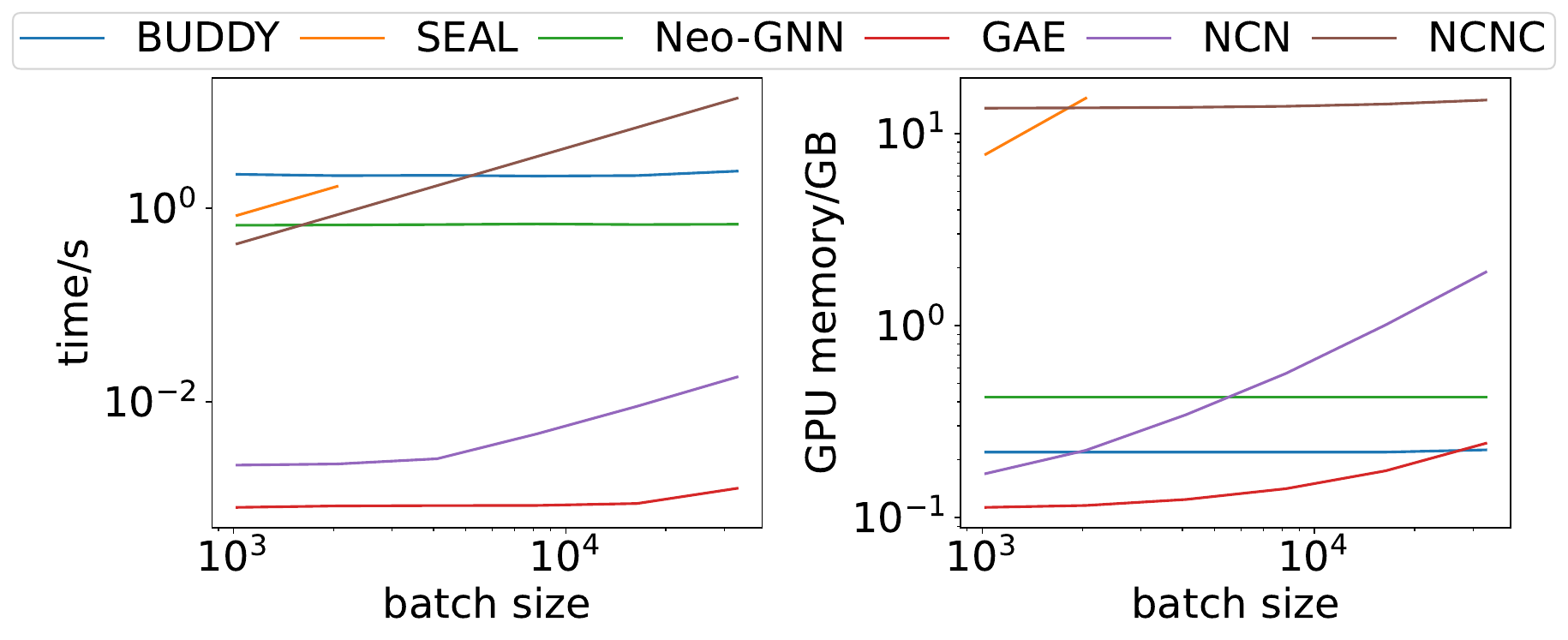}}
    \subfloat[Ogbl-citation2]{ \includegraphics[width=0.5\textwidth]{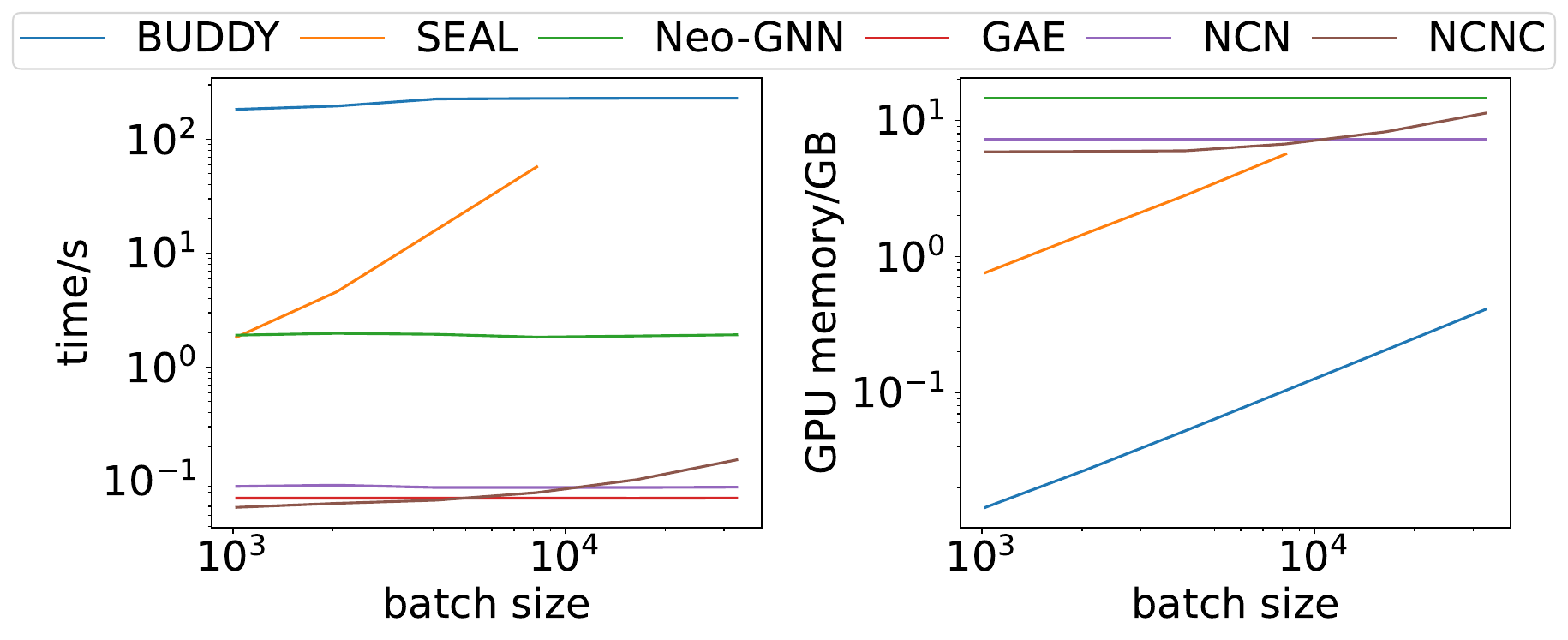}}
    \vskip -0.1in
    \caption{Inference time and GPU memory on datasets. The process we measure includes preprocessing, MPNN, and predicting one batch of test links. } 
    \label{fig:apptime}
    %\vskip -0.2in
\end{figure}
Let $t$ denote the number of target links, $n$ denote the number of nodes in the graph, and $d$ denote the maximum node degree. Existing models' time and space complexity can be expressed in $O(B+Ct)$ and $O(D+Et)$ respectively, where $B, C, D, E$ are irrelevant to $t$. $B,C,D,E$ of models are summarized in Table~\ref{tab:complexity}. The derivation of the complexity is as follows. As NCN, GAE, and GNN with separated structural features run MPNN on the original graph, they share similar $ndF+nF^2$ in $B$. Specifically, BUDDY~\citep{Gsketch} uses a simplified MPNN with $ndF$ in $B$. Moreover, Neo-GNN needs to precompute high order graph $A^l$, which takes $O(nd^l)$ time and space. BUDDY needs to hash each node and takes $O(nh)$ time and $O(nh')$ space. In contrast, $B$ of SEAL is $0$ as it does not run MPNN on the original graph. For each target link, vanilla GNN only needs to feed the feature vector to MLP for each link, so $C=F^2$. Besides GAE's operation, BUDDY further needs to hash the structure for structural features, whose complexity is complex but higher than $d$ per edge, and Neo-GNN computes pairwise feature with $O(d^l)$ complexity, where $l$ is the number of hop Neo-GNN consider. NCN needs to compute common neighbor: $O(d)$, pool node embeddings: $O(dF)$, and feed to MLP: $O(F^2)$. NCNC-$1$ runs NCN for each potential common neighbor: $O(F^2+d(dF+F^2))=O(d^2F+dF^2)$. Similarly, NCNC-$K$ runs $O(d)$ times NCNC-$(K\!-\!1)$, so its time complexity is $O(d^{K+1}F+d^KF^2)$. For each target link, SEAL segregates a subgraph of size $O(d^{l'})$ and runs MPNN on it, so $C=d^{l'}F^2+d^{l'+1}F$, where $l'$ is the number of hops of the subgraph. 

\section{Scalability Comparison on datasets}\label{app:time}
The time and memory consumption of models on different datasets are shown in Figure~\ref{fig:apptime}. On these datasets, we observe results similar to those on the ogbl-collab dataset in Section~\ref{sec:scalability}: NCN achieves similar computation overhead to GAE; NCNC usually scales better than Neo-GNN; SEAL's scalabilty is the worst. However, on the ogbl-citation2 dataset, SEAL has the lowest GPU memory consumption with small batch sizes, because the whole graph in ogbl-citation2 is large, on which MPNN is expensive, while SEAL only runs MPNN on small subgraphs sampled from the whole graph, leading to lower overhead.

\begin{table}[t]
    \centering
    \caption{Results on link prediction benchmarks. The format is average score $\pm$ standard deviation.  NCN+tricks means NCN with tricks of PLNLP. }\label{tab:add_results}
    \begin{tabular}{lcccc}
    \toprule
        ~ & \textbf{Collab} & \textbf{PPA} & \textbf{Citation2} & \textbf{DDI} \\ 
    \midrule
        Metric & Hits@50 & Hits@100 & MRR & Hits@20 \\
    \midrule
        \textbf{NCN} & $64.76\pm 0.87$ & $61.19\pm 0.85$ & $88.64\pm 0.14$ & $82.32\pm 6.10$ \\ 
        \textbf{NCNC} & $66.61\pm 0.71$ & $61.42\pm 0.73$ & $89.12\pm 0.40$ & $84.11\pm 3.67$ \\ 
        \textbf{Node2Vec} & $41.36\pm 0.69$ & $27.83\pm 2.02$ & $53.47\pm 0.12$ & $21.95\pm 1.58$ \\ 
        \textbf{DeepWalk} & $50.37 \pm 0.34$ & $28.88 \pm 1.53$ & $84.48\pm 0.30$ & $26.42 \pm 6.10$ \\ 
        \textbf{LINE} & $55.13\pm 1.35$ & $26.03\pm 2.55$ & $82.33\pm 0.52$ & $10.15\pm 1.69$ \\ 
        \textbf{PLNLP} & $70.59\pm 0.29$ & $32.38\pm 2.58$ & $84.92\pm 0.29$ & $90.88\pm 3.13$ \\ 
        \textbf{GIDN}  & $70.96\pm 0.55$&-&-&-\\
        \textbf{NCN+tricks} & $68.04\pm 0.42$ & - & - & $90.83\pm 2.83$ \\ 
    \bottomrule
    \end{tabular}
\end{table}
\section{Comparison with other link prediction models}\label{app:plnlp}
\paragraph{Node embedding methods}
The main advantage of GNN methods is that they keep permutation equivariance. In other words, these methods can give isomorphic links (links with the same structure) the same prediction. In contrast, node embedding methods, such as Node2Vec~\citep{Node2Vec}, LINE~\citep{Line}, and DeepWalk~\citep{Deepwalk}, will produce different results for isomorphic links, leading to potentially bad generalization.

We also compare our method with representative node embedding methods on ogb datasets in Table~\ref{tab:add_results}. NCN and NCNC outperform node embedding methods significantly on all datasets, indicating the advantages of MPNNs considering pairwise features for link prediction.

\paragraph{Other GNNs}

Instead of representations of pairwise relations, PLNLP~\citep{PLNLP} and GIDN~\citep{GIDN} boost GNNs on link prediction tasks by training tricks like loss function and data augmentation. These tricks are orthogonal to our model design.  In experiments (Table~\ref{tab:add_results}), compared with PLNLP, NCN achieves $89\%$ performance gain on ogbl-ppa and $20\%$ gain on average. As GIDN only conducts experiment on one dataset ogbl-collab, the comparison is not complete. Moreover, tricks of PLNLP can also boost our models. 
\begin{figure}[t]
    \centering
    \includegraphics[width=\textwidth]{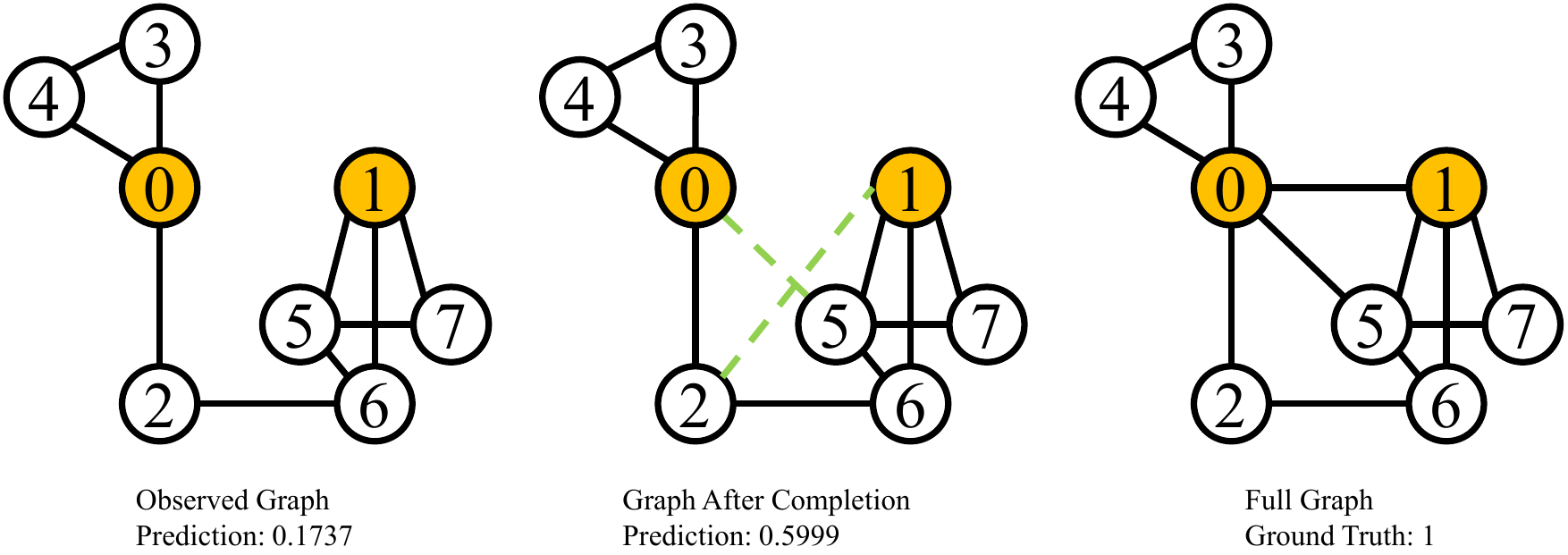}
    \caption{Visualization of how NCNC works. The example is from the Cora dataset, a citation graph. The target link is (0, 1). Models should produce high link existence probability. However, in the observed graph, link (0, 5) is missing, and (0,1) thus has no common neighbor. So NCN predicts that the link is not likely to exist. However, for NCNC, it first completes common neighbors (see green lines). Therefore, NCNC predicts that (0, 1) is more likely to exist. Note that NCNC completes common neighbors by probability, and we only plot completion with probability > 0.5 here. And the two completions are with about 0.95 probability. Though the common neighbor 2 completed by the model does not exist in the full graph, the full graph here only means a graph with all training, validation, and test edges, and the citation relation in the graph may still need to be completed.}
    \label{fig:CNCexample}
\end{figure}
\section{CNC Example}\label{app:cncexp}
Figure~\ref{fig:CNCexample} provides an example from Cora dataset on how CNC works.

{
\section{Ablation of MPNN}

Here we provide an ablation study on the MPNN used in NCN. The results are shown in Table~\ref{tab::MPNNabl}. The MPNN model includes GIN~\citep{HowPowerfulAreGNNs}, GraphSage~\citep{GraphSage}, MPNN with max aggregation, GCN~\citep{GCN}, and GAT~\citep{GAT}. Though the performance of NCN is sensitive to the MPNN model, NCN achieves performance gain with all GNNs compared with GraphAutoencoder (GAE). 

\begin{table}[!ht]
    \centering
    \caption{Ablation study on MPNN. }\label{tab::MPNNabl}
    \begin{tabular}{lcccccc}
    \toprule
        Dataset & Model & GIN & GraphSage & max & GCN & GAT \\ \midrule
        Cora&GAE & $70.45_{\pm 1.88}$ & $70.59_{\pm 1.70}$ & $61.63_{\pm 4.43}$ & $89.01_{\pm 1.32}$ & $83.36_{\pm 2.54}$ \\ 
        &NCN & $70.62_{\pm 1.68}$ & $70.94_{\pm 1.47}$ & $66.53_{\pm 2.27}$ & $89.05_{\pm 0.96}$ & $83.93_{\pm 2.03}$ \\ 
        Citeseer &GAE & $61.21_{\pm 1.18}$ & $61.23_{\pm 1.28}$ & $53.02_{\pm 3.75}$ & $91.78_{\pm 0.94}$ & $68.49_{\pm 2.75}$ \\ 
        &NCN & $61.58_{\pm 1.18}$ & $61.95_{\pm 1.05}$ & $53.40_{\pm 2.34}$ & $91.56_{\pm 1.43}$ & $69.27_{\pm 2.08}$ \\ 
        Pubmed&GAE & $59.00_{\pm 0.31}$ & $57.20_{\pm 1.37}$ & $55.08_{\pm 1.43}$ & $78.81_{\pm 1.64}$ & $74.44_{\pm 1.04}$ \\ 
        &NCN & $59.06_{\pm 0.49}$ & $58.06_{\pm 0.69}$ & $56.32_{\pm 0.77}$ & $79.05_{\pm 1.16}$ & $74.43_{\pm 0.81}$ \\ 
        collab&GAE & $38.94_{\pm 0.81}$ & $28.11_{\pm 0.26}$ & $27.08_{\pm 0.61}$ & $36.96_{\pm 0.95}$ & OOM \\ 
        &NCN & $64.38_{\pm 0.06}$ & $63.94_{\pm 0.43}$ & $64.19_{\pm 0.18}$ & $64.76_{\pm 0.87}$ & OOM \\ 
        ppa& GAE & $18.20_{\pm 0.45}$ & $11.79_{\pm 1.02}$ & $20.86_{\pm 0.81}$ & $19.49_{\pm 0.75}$ & OOM \\ 
        & NCN & $47.94_{\pm 0.89}$ & $56.41_{\pm 0.65}$ & $57.31_{\pm 0.30}$ & $61.19_{\pm 0.85}$ & OOM \\ 
    \bottomrule
    \end{tabular}
\end{table}

\section{Choice of Metrics}
We test our model in different metrics. The results are shown in Table~\ref{tab::metricabl}. In total, NCN achieves 11 best score (in bold), NCNC achieves 22 best score, and our strongest baseline achieves 9 best score. Therefore, our NCN and NCNC still outperforms baselines in different metrics. 
\begin{table}[h]
    \caption{Models' performance with various metrics. BUDDY is our strongest baseline. Blanks mean unfinished experiments due to time constraints. }\label{tab::metricabl}
    \centering
    \begin{small}
    \setlength{\tabcolsep}{0.3mm}\begin{tabular}{llccccccc}
    \toprule
        ~ & ~ & Cora & Citeseer & Pubmed & Collab & PPA & Citation2 & DDI \\ 
    \midrule
        hit@1 & NCN & $\mathbf{16.24_{\pm 14.18}}$ & $29.32_{\pm 18.19}$ & $7.03_{\pm 6.10}$ & $4.94_{\pm 2.95}$ & $5.91_{\pm 4.11}$ & $83.79_{\pm 0.06}$ & $0.24_{\pm 0.11}$ \\ 
        ~ & NCNC & $10.90_{\pm 11.40}$ & $\mathbf{32.45_{\pm 17.01}}$ & $\mathbf{8.57_{\pm 6.76}}$ & $9.82_{\pm 2.49}$ & $\mathbf{7.78_{\pm 0.63}}$ & ~ & $0.16_{\pm 0.07}$ \\ 
        ~ & BUDDY & $11.74_{\pm 5.77}$ & $20.87_{\pm 12.22}$ & $2.97_{\pm 2.02}$ & $\mathbf{10.71_{\pm 0.64}}$ & $2.29_{\pm 1.26}$ & ~ & $\mathbf{2.40_{\pm 4.81}}$ \\ 
    \midrule
        hit@3 & NCN & $29.52_{\pm 13.79}$ & $49.98_{\pm 14.49}$ & $\mathbf{19.16_{\pm 4.39}}$ & $11.07_{\pm 6.32}$ & $15.32_{\pm 3.31}$ & $92.41_{\pm 0.06}$ & $1.54_{\pm 3.43}$ \\ 
        ~ & NCNC & $25.04_{\pm 11.40}$ & $\mathbf{50.49_{\pm 12.01}}$ & $17.58_{\pm 6.57}$ & $\mathbf{21.07_{\pm 5.46}}$ & $\mathbf{16.58_{\pm 0.60}}$ & ~ & $0.59_{\pm 0.42}$ \\ 
        ~ & BUDDY & $\mathbf{32.67_{\pm 10.10}}$ & $41.16_{\pm 9.12}$ & $10.41_{\pm 4.16}$ & $16.25_{\pm 1.59}$ & $7.75_{\pm 0.48}$ & ~ & $\mathbf{10.84_{\pm 7.55}}$ \\ 
    \midrule
        hit@10 & NCN & $\mathbf{55.87_{\pm 4.40}}$ & $\mathbf{69.68_{\pm 3.05}}$ & $\mathbf{34.61_{\pm 5.02}}$ & $43.51_{\pm 1.84}$ & $25.76_{\pm 3.65}$ & $96.50_{\pm 0.06}$ & $40.04_{\pm 19.59}$ \\ 
        ~ & NCNC & $53.78_{\pm 7.33}$ & $69.59_{\pm 4.48}$ & $34.29_{\pm 4.43}$ & $43.22_{\pm 6.19}$ & $\mathbf{26.67_{\pm 1.51}}$ & ~ & $45.64_{\pm 14.12}$ \\ 
        ~ & BUDDY & $50.98_{\pm 3.46}$ & $67.05_{\pm 2.83}$ & $23.92_{\pm 5.01}$ & $\mathbf{53.11_{\pm 0.86}}$ & $17.41_{\pm 0.06}$ & ~ & $\mathbf{52.70_{\pm 7.70}}$ \\ 
    \midrule
        hit@20 & NCN & $\mathbf{68.31_{\pm 3.00}}$ & $78.02_{\pm 1.99}$ & $50.94_{\pm 3.11}$ & $55.87_{\pm 0.36}$ & $\mathbf{37.57_{\pm 1.98}}$ & $97.87_{\pm 0.04}$ & $82.55_{\pm 4.08}$ \\ 
        ~ & NCNC & $67.10_{\pm 2.96}$ & $\mathbf{79.05_{\pm 2.68}}$ & $\mathbf{51.42_{\pm 3.81}}$ & $57.83_{\pm 3.14}$ & $35.00_{\pm 2.22}$ & ~ & $\mathbf{83.92_{\pm 3.25}}$ \\ 
        ~ & BUDDY & $61.92_{\pm 2.67}$ & $76.15_{\pm 3.31}$ & $34.75_{\pm 5.12}$ & $\mathbf{59.06_{\pm 0.57}}$ & $27.28_{\pm 0.52}$ & ~ & $78.14_{\pm 4.23}$ \\ 
    \midrule
        hit@50 & NCN & $80.85_{\pm 1.12}$ & $86.33_{\pm 1.55}$ & $67.77_{\pm 1.91}$ & $64.45_{\pm 0.35}$ & $\mathbf{51.54_{\pm 1.48}}$ & $99.01_{\pm 0.02}$ & $94.17_{\pm 0.36}$ \\ 
        ~ & NCNC & $\mathbf{81.36_{\pm 1.86}}$ & $\mathbf{88.60_{\pm 1.51}}$ & $\mathbf{69.25_{\pm 2.87}}$ & $\mathbf{66.88_{\pm 0.66}}$ & $48.66_{\pm 0.18}$ & ~ & $\mathbf{94.85_{\pm 0.56}}$ \\ 
        ~ & BUDDY & $76.64_{\pm 2.45}$ & $85.46_{\pm 2.17}$ & $55.75_{\pm 3.38}$ & $66.09_{\pm 0.48}$ & $39.99_{\pm 0.02}$ & ~ & $92.17_{\pm 0.95}$ \\ 
    \midrule
        hit@100 & NCN & $\mathbf{89.14_{\pm 1.04}}$ & $91.82_{\pm 1.14}$ & $79.56_{\pm 1.11}$ & $67.25_{\pm 0.15}$ & $61.25_{\pm 0.61}$ & $99.51_{\pm 0.02}$ & $97.09_{\pm 0.43}$ \\ 
        ~ & NCNC & $89.05_{\pm 1.24}$ & $\mathbf{93.13_{\pm 1.13}}$ & $\mathbf{81.18_{\pm 1.24}}$ & $\mathbf{71.96_{\pm 0.14}}$ & $\mathbf{62.02_{\pm 0.74}}$ & ~ & $\mathbf{97.60_{\pm 0.22}}$ \\ 
        ~ & BUDDY & $84.82_{\pm 1.96}$ & $91.48_{\pm 1.15}$ & $70.92_{\pm 2.08}$ & $70.53_{\pm 0.17}$ & $48.07_{\pm 0.05}$ & ~ & $95.38_{\pm 0.65}$ \\ 
    \midrule
        mrr & NCN & $\mathbf{29.20_{\pm 13.59}}$ & $43.93_{\pm 12.87}$ & $\mathbf{17.44_{\pm 3.40}}$ & $13.76_{\pm 2.49}$ & $13.48_{\pm 2.83}$ & $88.62_{\pm 0.05}$ & $5.48_{\pm 1.23}$ \\ 
        ~ & NCNC & $23.55_{\pm 9.67}$ & $\mathbf{45.64_{\pm 11.78}}$ & $15.63_{\pm 4.13}$ & $17.68_{\pm 2.70}$ & $\mathbf{14.37_{\pm 0.06}}$ & ~ & $8.61_{\pm 1.37}$ \\ 
        ~ & BUDDY & $27.28_{\pm 4.71}$ & $35.77_{\pm 9.59}$ & $10.79_{\pm 2.81}$ & $\mathbf{18.97_{\pm 0.50}}$ & $7.47_{\pm 0.02}$ & ~ & $\mathbf{13.53_{\pm 6.07}}$ \\ 
    \bottomrule
    \end{tabular}
    \end{small}
\end{table}

}
\end{document}